\documentclass[journal]{IEEEtran}

\usepackage{tabularx,booktabs,multirow}
\usepackage[table]{xcolor}

\usepackage{ifpdf}
  \ifpdf
  \else
  \fi

\usepackage{cite}

\ifCLASSINFOpdf
  \usepackage[pdftex]{graphicx}
\else
\fi

\usepackage{amsmath}
\usepackage{soul, color, xcolor}

\soulregister{\cite}7
\soulregister{\ref}7

\usepackage{algorithmic}
\usepackage{algorithm}

\usepackage{array}

\ifCLASSOPTIONcompsoc
  \usepackage[caption=false,font=normalsize,labelfont=sf,textfont=sf]{subfig}
\else
  \usepackage[caption=false,font=footnotesize]{subfig}
\fi

\usepackage{stfloats}
\ifCLASSOPTIONcaptionsoff
  \usepackage[nomarkers]{endfloat}
 \let\MYoriglatexcaption\caption
 \renewcommand{\caption}[2][\relax]{\MYoriglatexcaption[#2]{#2}}
\fi

\usepackage{url}

\begin{document}

\title{Asymmetric Modality Translation For Face Presentation Attack Detection}

\author{Zhi Li,
        Haoliang Li,
        Xin Luo,
        Yongjian Hu,
        Kwok-Yan Lam,
        and Alex C. Kot~\IEEEmembership{Fellow,~IEEE}
}
\maketitle
\begin{abstract}
Face presentation attack detection (PAD) is an essential measure to protect face recognition systems from being spoofed by malicious users and has attracted great attention from both academia and industry. Although most of the existing methods can achieve desired performance to some extent, the generalization issue of face presentation attack detection under cross-domain settings (e.g., the setting of unseen attacks and varying illumination) remains to be solved. In this paper, we propose a novel framework based on asymmetric modality translation for face presentation attack detection in bi-modality scenarios. Under the framework, we establish connections between two modality images of genuine faces. Specifically, a novel modality fusion scheme is presented that the image of one modality is translated to the other one through an asymmetric modality translator, then fused with its corresponding paired image. The fusion result is fed as the input to a discriminator for inference. The training of the translator is supervised by an asymmetric modality translation loss. Besides, an illumination normalization module based on Pattern of Local Gravitational Force (PLGF) representation is used to reduce the impact of illumination variation. We conduct extensive experiments on three public datasets, which validate that our method is effective in detecting various types of attacks and achieves state-of-the-art performance under different evaluation protocols.
\end{abstract}

\begin{IEEEkeywords}
Face Presentation Attack Detection, Asymmetric Modality Translation, Modality Fusion, Unseen Attack, Illumination Normalization.
\end{IEEEkeywords}

\section{Introduction}
\IEEEPARstart{F}{ace} recognition has become one of the most predominant methods for automatic identity verification due to its high accuracy and convenience. It has been widely used in diverse application scenarios, ranging from attendance registration and intelligent device unlocks to security-critical ones such as door access control, border control, and cashless payment. Compared to other biometric-based methods, face recognition is contactless and silent during the information acquisition process. Despite its ease of use, face images could be inadvertently disclosed and easily captured by malicious users, especially with the widespread use of social media. Using leaked faces of genuine users, the malicious users can create diverse spoofing artifacts such as printed photos, replayed videos, masks made of different materials, and even wax figures. To bypass the identity verification mechanism and be recognized as genuine users, these artifacts could thereby be utilized to conduct presentation attacks on victim systems by malicious users. Presentation attack, a.k.a. spoofing attack, is a physical-layer attack that works on the sensing process of identity verification systems. Because of its non-intrusive characteristic and ease of deployment, it becomes one of the main threats to face recognition systems and hinders their application. 

\begin{figure}[tbp]
\centering
\includegraphics[width=0.45 \textwidth]{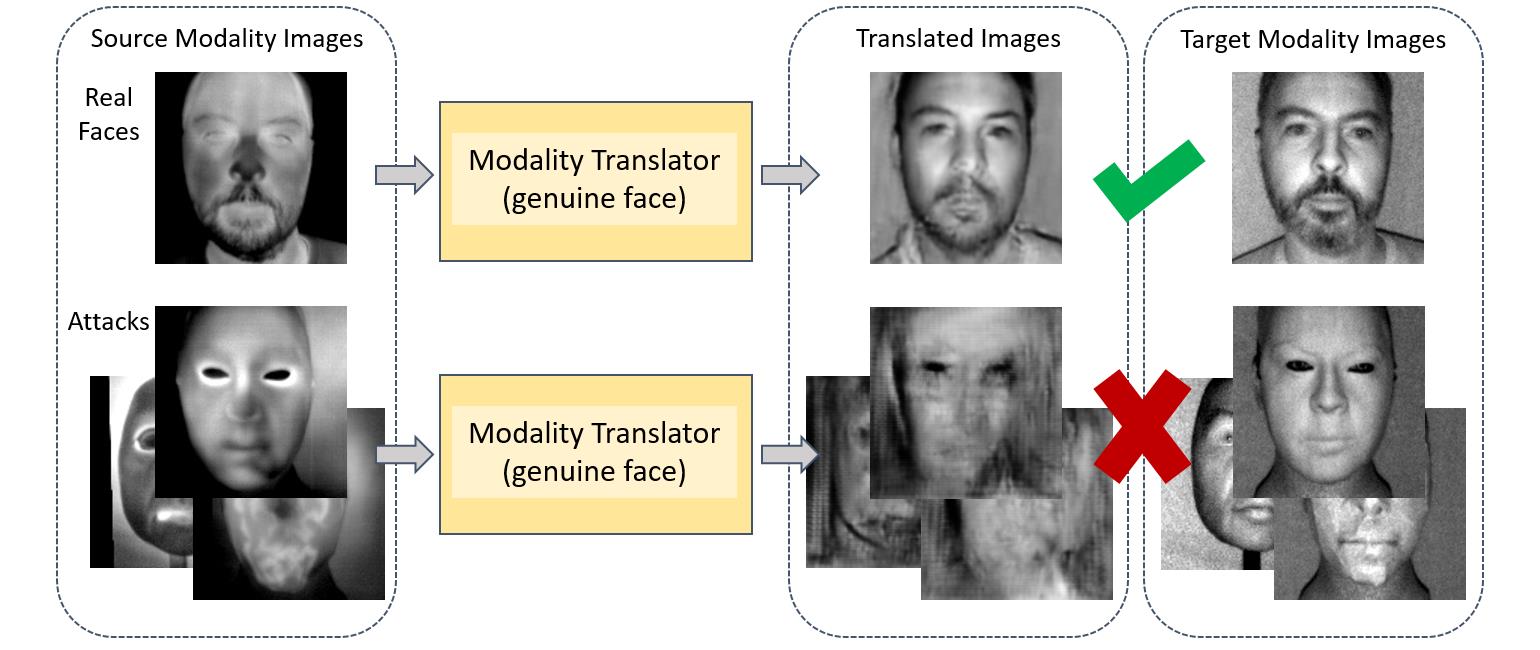}
\caption{Illustration of asymmetric modality translation. The modality translator can successfully translate genuine face images to the target modality while fails for attacks.}
\label{fig:figure_01}
\end{figure}

Face presentation attack detection (PAD) is an essential measure to protect face recognition systems from being spoofed by malicious users, which has attracted great attention from both academia and industry. Face PAD, in essence, is a task of presented ``face'' perception and discrimination. A diversity of sensor devices have been used for information acquisition in research. Due to the wide application of visible light (VIS) cameras on multifarious daily-used electronic devices, VIS-based methods naturally became the mainstream of face PAD research at the early stage. Although deep learning (DL)-based methods have achieved good performance in intra-domain tests, most of them can not generalize well to domain shift caused by factors such as varying illumination conditions and unseen attacks. 

Thanks to the development of sensor technology and the manufacturing process, the price of multi-modality sensors continues to drop, which makes them affordable to be equipped on up-to-date mobile devices. Recently, multi-modality-based face PAD methods have shown promising results and become a research hot spot. Existing multi-modality-based face PAD methods improved the performance to some extent, but like single-modality ones, the generalization problems to unseen attack and the variation of environment illumination still remain to be solved.

In this work, we tackle the generalization problems in face PAD under bi-modality scenarios and propose a method that generalizes better to unseen attack and illumination variations. Different from existing works, we explicitly establish the connection between different modality images of genuine faces via asymmetric modality translation as shown in Fig. 1, and use it as the core to build a bi-modality face PAD framework with higher accuracy and robustness to illumination variations and unseen attacks. Our main contribution in this work can be summarized as below:

\begin{itemize}
\item We propose a novel framework based on asymmetric modality translation for face presentation attack detection under bi-modality scenarios.

\item We propose an asymmetric modality translation loss to supervise the training of the translator at both latent and pixel-level and design an illumination normalization module based on PLGF to reduce the effect of illumination variations on sensitive modalities.

\item We conduct extensive experiments to verify the effectiveness of our proposed method. The results show that our method applies to different modality settings and achieves state-of-the-art performance under grand-test, cross-illumination, and unseen-attack evaluation protocols.
\end{itemize}

\section{Related Works}
In the literature, a diversity of sensor devices are used for information acquisition. In addition to the most commonly used VIS camera, NIR \cite{SPRINGER15_MSSPOOF, CVPR19_CASIA_SURF, WACV21_CeFA, TIFS20_MCCNN, TBIOM20_MCDPB}, depth \cite{CVPR19_CASIA_SURF, WACV21_CeFA, TIFS20_MCCNN, TBIOM20_MCDPB}, thermal \cite{TIFS20_MCCNN, TBIOM20_MCDPB}, short-wave infrared (SWIR)\cite{TBIOM20_MCDPB}, ultraviolet (UV) \cite{BIOSIG20_ULTRA}, light-field \cite{TIP15_LFC, TIFS18_LFC} and dual pixel \cite{TIFS21_DPC} cameras have also been used for face anti-spoofing. According to the sensor devices used at the test phase, face PAD methods can be broadly categorized into single-modality and multi-modality-based. Although single VIS-modality-based methods have been extensively studied for many years, the generalization problem still remains to be solved. Recently, multi-modality-based methods \cite{CVPR19_CASIA_SURF, TIFS20_MCCNN, TBIOM20_MCDPB} have achieved better performance with the help of richer information gathered from different modalities \cite{SPRINGER15_MSSPOOF, WACV21_CeFA, CVPR19_CASIA_SURF, TIFS20_MCCNN, TBIOM20_MCDPB}.

In the following content of this section, the focus of our review is on the methods based on single VIS-modality and multi-modality, which are most relevant to our work.  

\subsection{Single VIS-modality-based Face Presentation Attacks Detection\label{section:single-vis-pad}}

Due to the wide application of VIS cameras on multifarious daily-used electronic devices, VIS-based methods naturally became the mainstream of face PAD research at the early stage. Various types of hand-crafted features based on textures \cite{ICIP15_LBP, BTAS16_HARALICK}, image quality \cite{TIP14_IQA, TIFS15_IDA} and liveness signals \cite{ECCV16_RPPG, TIFS21_MCRPPG} have been designed by domain experts to discriminate genuine and spoofing face images. The designs of these features are usually based on some prior knowledge about specific types of attacks. Although hand-crafted feature-based methods have strong interpretability, they can hardly take care of various types of attacks and fail when the presupposed cue does not exist.

With the advent of the deep learning era, CNN-based methods demonstrate remarkable advantages under intra-domain evaluation settings, benefiting from large amounts of labeled data. However, the performances of plain data-driven methods are limited by the diversity of training data \cite{FG17_OULUNPU}. These methods can not generalize well under domain shift caused by the variations of illumination conditions and capture devices. To tackle the generalization problem, \cite{TIFS18_UDAFAS, CVPR20_MDDRL, CVPR20_SSDG, TIFS20_UADA, TIP21_PTL} propose methods to learn more generalized features by leveraging the advances of domain generalization and adaptation techniques. \cite{TIFS21_DRLFAS} models the behavior of exploring spoofing cues from sub-patches of images with deep reinforcement learning. \cite{CVPR20_CDCN, TPAMI21_NASFAS} propose network architectures based on central difference convolution (CDC) layers for fine-grained feature learning and automatically search the optimal parameters via network architecture search (NAS) techniques. 

Besides the methods based on learning and optimization theories, there are some works stepping on the roads from specific physical meanings. \cite{TIFS18_GDFR, TIFS19_MOTIONBLUR, CVPR20_DSGTD, TBIOM21_ABSTMCN} design methods to extract auxiliary information from temporal spaces. \cite{ECCV18_Denoising, ECCV20_ODPT} tackle the problem of face PAD via modeling the noises or traces of spoofing face images. \cite{TIFS20_SRA} combines the intrinsic image decomposition with deep learning to improve the generalization ability. \cite{ECCV20_BCN} propose networks to capture intrinsic material-based patterns. In addition to binary supervision, pseudo-depth maps\cite{CVPR18_BA}, pseudo-reflection maps \cite{ECCV20_CELEBASPOOF}, texture maps \cite{ECCV20_LBP}, binary masks\cite{ICB19_DPB, TIFS20_SUN} and NIR images\cite{TIFS21_FASACMT} have been used as auxiliary information for the supervision of VIS-models at the training phase. Recently, \cite{TBIO21_Revisiting} propose a pyramid supervision to learn both local details and global semantics. Along with the up-gradation of face PAD methods, new types of attacks are emerging and varying ceaselessly. \cite{CVPR19_DTL, ICASSP20_HYPER, TIFS21_OCR_MCCNN} address face PAD problem under a more realistic scenario and design methods against attacks that unseen at training phase. Although VIS-based face anti-spoofing has been extensively studied for years, the generalization problem to unseen attack and illumination variation is still an open issue of great challenge.

\subsection{Multi-modality-based Face Presentation Attack Detection}
Compared to the single VIS-based face PAD problem, multi-modality face PAD is still under-explored. This is mainly due to the cost of using additional sensor devices. Recently, the falling price of multi-modality sensors makes them affordable to be equipped on the latest mobile devices. Several large-scale datasets \cite{CVPR19_CASIA_SURF, TBIOM_CASIA_SURF, TIFS20_MCCNN, TBIOM20_MCDPB, WACV21_CeFA} containing paired multi-modality data have been released in the past two years to support the study and evaluation of multi-modality face PAD methods.

\cite{CVPR19_CASIA_SURF} propose a large-scale dataset of VIS, NIR, depth modalities, and a method based on ResNet18 \cite{CVPR16_RESNET} with squeeze-and-excitation (SE) fusion \cite{CVPR18_SE} to detect printed photo attacks in multi-modality scenarios. Based on \cite{CVPR19_CASIA_SURF}, \cite{TBIOM_CASIA_SURF} improves the performance by extending the network architecture and adjusting the data augmentation and fusion scheme. At the same period, \cite{TIFS20_MCCNN} introduces a dataset, which contains VIS, NIR, thermal and depth modality data of genuine faces, and seven types of attacks. In addition, a Multi-Channel Convolutional Neural Network (MC-CNN), based on LightCNN\cite{TIFS18_LIGHTCNN}, is proposed for multi-modality-based face PAD. \cite{TBIOM20_MCDPB} extends the WMCA dataset with SWIR modality data and proposes the MC-DeepPixBiS method, which fuses aligned images of different modalities at the input level. The network training is supervised with both pixel-wise binary and binary labels by drawing lessons from the advances in VIS-based face PAD method \cite{ICB19_DPB}. Similarly, \cite{CVPRW20_MCCDCN} proposes a method based on CDC \cite{CVPR20_CDCN} for multi-modality face PAD task. Recently, \cite{WACV21_CeFA} collects a dataset that is specifically for cross-ethnic face PAD research. Besides, there are some methods using additional modalities at the training phase as mentioned in Section \ref{section:single-vis-pad}. Although \cite{CVPR18_BA, ECCV20_CELEBASPOOF, TIFS21_FASACMT} use pseudo-depth maps, pseudo-reflection maps, or NIR images as auxiliary information to assist model training, they only use VIS images for inference at the test phase. Unlike them, our method aims at improving the performance for detecting various types of attacks under multi-modality scenarios that is the same as \cite{TIFS20_MCCNN, TBIOM20_MCDPB, CVPR19_CASIA_SURF, TBIOM_CASIA_SURF, CVPRW20_MCCDCN}. 

The majority of existing multi-modality-based face PAD methods are simple extensions from VIS-based ones. The intrinsic characteristics of the face PAD task and the relationships between different modalities are scarcely considered. Compared to other multi-modality-based face PAD methods, our method focuses on the modality translation of genuine face images, and leverages the consistency between the translated images and their ground-truth pairs for discrimination, which mitigates the overfit to seen types of attacks to some extent. Besides, our method includes an illumination normalization module, which improves the robustness to the variation of illumination conditions. To verify the effectiveness and advantages of the method itself, we focus on comparing it with multi-modality-based methods under fair evaluation settings in Section \ref{section:experimental-results}.
\section{Proposed Method}
In this section, we present the motivation and design details of our proposed framework in concrete.
\begin{figure*}[htbp]
\centering
\includegraphics[width=0.8\textwidth]{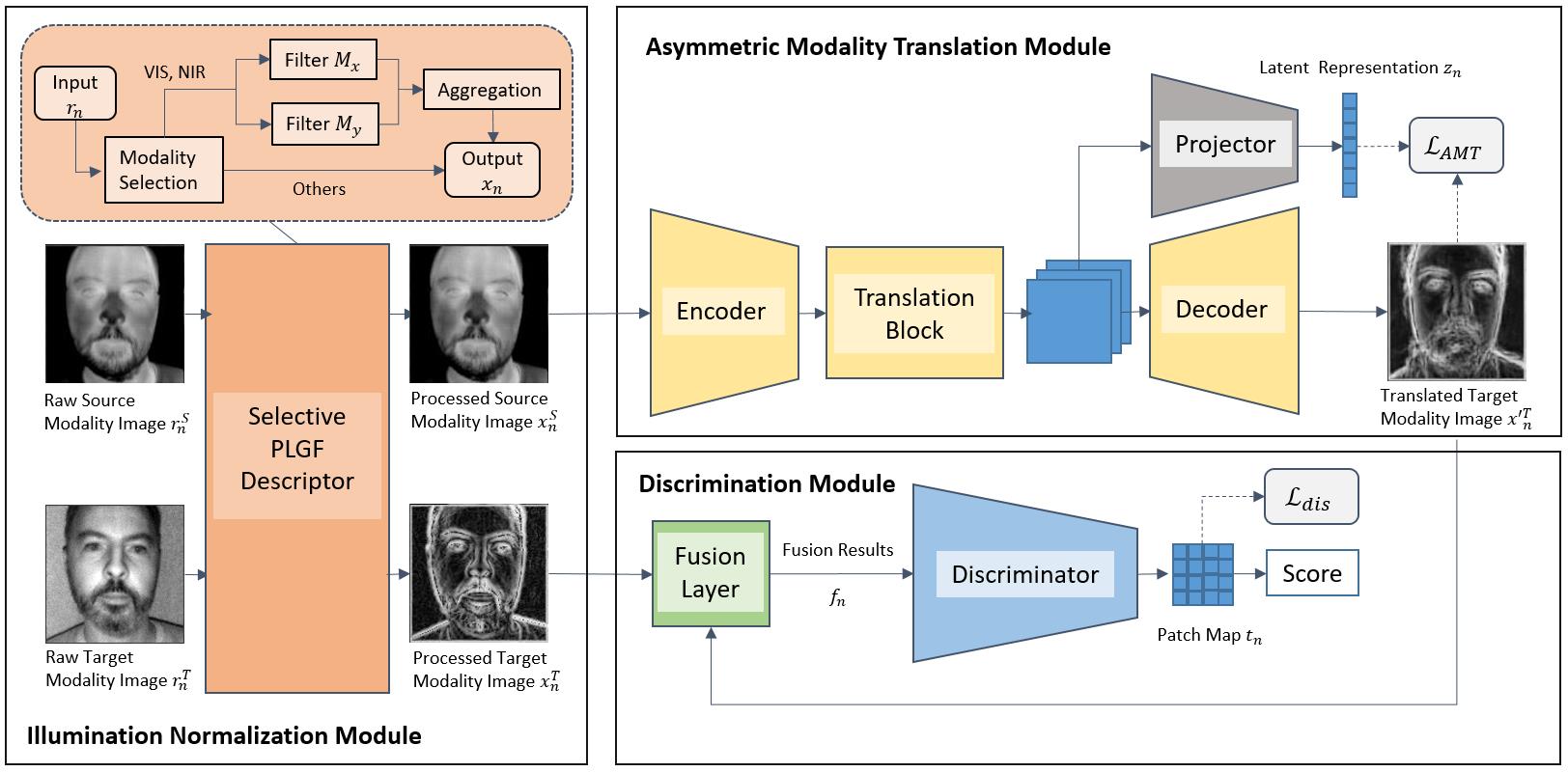}
\caption{The proposed framework consists of three modules. The Illumination Normalization (IN) Module selectively reduces the impact of illumination variations on sensitive modalities. The Asymmetric Modality Translation (AMT) Module translates the source modality image to the target modality. The Discrimination (DC) Module fuses the translated image with the ground-truth target modality image as the input for inference.}
\label{fig:figure_02}
\end{figure*}
For most multi-modality-based face PAD methods, images of different modalities are processed individually in different branches or naively stacked together at the input level where the relationships between paired images of different modalities are scarcely considered. In this work, we aim to formulate the relationship between different modalities and leverage it to help detect face spoofing attacks. We find that the connection between VIS and NIR images can be established based on Illumination-Reflectance Model (IRM) \cite{BOOK86_IRM}. According to IRM, each pixel $I(i,j)$ of an image is represented as the product of the reﬂectance $R(i,j)$ and the illumination components $L(i,j)$. VIS image and NIR image thus can be represented as
\begin{align}
\label{eq:equation_01}
I_{V}(i,j) &= R_{V}(i,j)\times L_{V}(i,j),\\
\label{eq:equation_02}
I_{I}(i,j) &= R_{I}(i,j)\times L_{I}(i,j). 
\end{align}

The simultaneous equation of Eq. (\ref{eq:equation_01}) and (\ref{eq:equation_02}) describes the relationship between paired images of VIS and NIR modality as

\begin{equation}
\begin{aligned}
I_{V}(i,j) = \frac {R_{V}(i,j)}{R_{I}(i,j)}\times \frac{L_{V}(i,j)}{L_{I}(i,j)}\times I_{I}(i,j).
\end{aligned}
\label{eq:equation_03}
\end{equation}
$R_{V}(i,j)$ and $R_{I}(i,j)$ of the first term are reflectance components of the object. While $L_{V}(i,j)$ and $L_{I}(i,j)$ of the second term are illumination components that depend on the external illumination condition during the image capturing process. Since the illumination components are irrelevant factors for the discrimination of genuine face and spoof artifacts, we did further analysis under a fixed illumination condition. We observe that, at the same illumination condition, the transform function from NIR modality $I_{I}(i,j)$ to VIS modality $I_{V}(i,j)$ is related to the reflectances of the object to VIS and NIR light. Since spoof artifacts such as printed photos, digital screens, and 3D masks are usually made of materials that are different from human skins, the transform functions for attacks should not be the same as the ones for genuine face images. And the transform function for genuine faces can not translate the spoofing faces correctly.

Motivated by the example of VIS-NIR modality, we aim to construct the transform function $T_{G}$ in the case of general bi-modality, which can successfully transform genuine face images from the source to the target modality but fails for attack ones as represented in Eq.(\ref{eq:equation_04}). ${x}_{g}^{S}$, ${x}_{g}^{T}$ and ${x'}_{g}^{T}$ denote source modality, target modality and translated images of genuine faces. Similarly, ${x}_{a}^{S}$, ${x}_{a}^{T}$ and ${x'}_{a}^{T}$ denote attack ones. The illustration is shown in Fig. 1.
\begin{equation}
\begin{aligned}
{x'}_{g}^{T} = T_{G}({x}_{g}^{S}) = {x}_{g}^{T},
\\
{x'}_{a}^{T} = T_{G}({x}_{a}^{S}) \neq {x}_{a}^{T}
.\\
\end{aligned}
\label{eq:equation_04}
\end{equation}

In this work, we leverage such discrepancy as an effective cue for discriminating various spoofing faces from genuine faces. Due to the different sensing and imaging principles of different modalities, it's difficult to formularize the general transform function directly. Instead, in our proposed framework, we implement the asymmetric transform function $T_{G}$ by convolutional neural networks (CNN) to make it applicable to different modalities.


\subsection{Framework of Proposed Method}
Based on the analysis mentioned above, we start to introduce the framework of our proposed method. As is shown in Fig. \ref{fig:figure_02}, the framework consists of three modules: Asymmetric Modality Translation (AMT) Module, Illumination Normalization (IN) Module, and Discrimination (DC) Module. 

This framework works at bi-modality scenarios, where paired images are synchronously captured using camera sensors of two different modalities. We set one modality as the source and the other as the target. After a series of preprocessing such as face detection-alignment-cropping-resizing, face images firstly pass through the IN module for illumination normalization. The source modality image is fed into the AMT module and translated to the target modality in an asymmetric way. The DC module fuses the translated image from the source modality with the ground-truth target modality image captured by the camera sensor in image space and distinguishes attack samples from genuine ones based on the output patch anomaly map. The working mechanisms and design details of each module will be introduced in the following content of this section.

\subsection{Asymmetric Modality Translation}
As elaborated at the beginning of this section, the asymmetric transform function is the core of our method. We expect that the function can successfully transform genuine face images from the source to the target modality but fails for attack ones. With this objective, we seek ways to construct our expected modality translator. Image-to-image (I2I) translation methods\cite{TMM20_I2I1, TMM20_I2I2, TMM21_AgeGAN, TMM21_I2I1, TMM21_I2I2, TMM21_I2I3} based on convolutional neural networks (CNN) have recently achieved promising results, which can automatically translate an image from the source space to the specific target space. Leveraging advances of them, the expected transformation function is implemented with a CNN-based translator. 

As illustrated in Fig. \ref{fig:figure_02}, the main branch of the AMT module is a translator of autoencoder architecture, which is similar to \cite{CVPR17_PIX2PIX} and commonly used in general I2I translation tasks. The encoder network consists of 3 convolution layers with instance normalization and ReLU as activation function, which maps data samples from the pixel space to a latent feature space. The translation block is implemented by residual blocks as proposed in ResNet \cite{CVPR16_RESNET}, which is used for modality translation at the latent space. The architecture of the decoder is symmetric to the encoder, which transforms data samples to the pixel space of the target modality. In addition to the main branch, we use an additional projector to assist the training of the translator, which will be discarded at the inference phase. The projector consists of two convolution layers that map the data to a lower-dimensional space.

To make the translator meet our goals, we propose an asymmetric modality translation loss to supervise the training of the translator at both pixel and latent feature levels. In the pixel space, the outputs ${x'}_{n}^{T}$ of the decoder are asymmetrically supervised by the ground-truth target modality images $x_{n}^{T}$ and class labels $y_{n}$. As is represented in Eq. (\ref{eq:equation_05}), the loss term $\mathcal{L}_{pixel}$ is controlled by a binary factor corresponding to class labels $y_{n}$, which enforces the training of the translator focus on genuine face images only and thereby achieves the goal of asymmetric translation. $N$, $N_{g}$, and $n$ are the total number of samples, the number of genuine samples, and the sample index. For simplification, Eq. (\ref{eq:equation_05}) can be rewritten into Eq. (\ref{eq:equation_06}), where $G$ is the set of indices of genuine face samples.
\begin{equation}
\begin{aligned}
\mathcal{L}_{pixel} & = \frac{1}{N_{g}} \sum_{n=1}^{N} (1 - y_{n}) \parallel {x'}_{n}^{T} - x_{n}^T  \parallel, \\
y_{n} & =\left\{
\begin{aligned}
0, &     & genuine,\\
1, &     & attack.
\end{aligned}
\right.
\end{aligned}
\label{eq:equation_05}
\end{equation}

\begin{equation}
\begin{aligned}
\mathcal{L}_{pixel} & = \frac{1}{N_{g}} \sum_{n \in G}
\parallel {x'}_{n}^{T} - x_{n}^{T}  \parallel.
\end{aligned}
\label{eq:equation_06}
\end{equation}

In the latent space, the discrepancy between genuine and attack samples is explicitly enlarged with a loss term adapted from the supervised contrastive loss \cite{NIPS20_SuperCon}. If we use $G$ and $A$ to denote the sets of indices of genuine and attack samples, respectively, the sizes of two sets can be represented by $|G|$ and $|A|$. Each genuine data sample can be paired with other genuine ones to generate $|G|-1$ positive pairs and with attacks to generate $|A|$ negative pairs. In our expected latent space, we wish the similarity of positives pairs is greater than negative pairs to ensure the consistency of genuine samples and the discrepancy between genuine and attack samples. The disparity to expectation can be measured by our asymmetric supervised contrastive loss. For each genuine sample $z_{n}$, the loss value $\mathcal{L}_{latent, n}$ can be represented as
\begin{equation}
\mathcal{L}_{latent, n}  = \frac{1}{|G|-1} \sum_{{g \in G }, {g \ne n}} - \log \frac{\exp(z_{n} \cdot z_{g}/\tau)}{\mathop{\sum}\limits_{a \in A} \exp(z_{n} \cdot z_{a}/\tau)}.
\label{eq:equation_07}
\end{equation}
$z_{n}$, $z_{g}$ are latent features of genuine samples with index $n$ and $g$, $z_{a}$ is latent feature of attack samples with index $a$, temperature $\tau$ is constant hyperparameter. The inner-product of two samples in latent space is used to measure the similarity. The relativeness is formulated by contrasting each positive pair with the sum of all negative pairs. Since the range of logarithmic term is $(-\infty, \infty)$, we truncate it at a fixed value $c$ to make it stable and compatible to ${L}_{pixel}$.
The loss over all genuine samples is represented as
\begin{footnotesize}
\begin{equation}
\mathcal{L}_{latent}  = \frac{1}{|G|-1} \sum_{n \in G} \sum_{{g \in G }, {g \ne n}} \max (- \log \frac{\exp(z_{n} \cdot z_{g}/\tau)}{\mathop{\sum}\limits_{a \in A} \exp(z_{n} \cdot z_{a}/\tau)}, c).
\label{eq:equation_08}
\end{equation}
\end{footnotesize}
By combining the loss term in both pixel and latent space, we get the fully asymmetric modality translation loss function in Eq. (\ref{eq:equation_09}). Supervised by which the CNN-based translator is trained to meet our expected functionalities.

\begin{equation}
\mathcal{L}_{AMT}  = \lambda_{1} \cdot \mathcal{L}_{pixel} + \lambda_{2} \cdot \mathcal{L}_{latent}.
\label{eq:equation_09}
\end{equation}
\subsection{Modality Fusion and Discrimination}
\begin{figure}[tbp]
\centering
\includegraphics[width=0.48\textwidth]{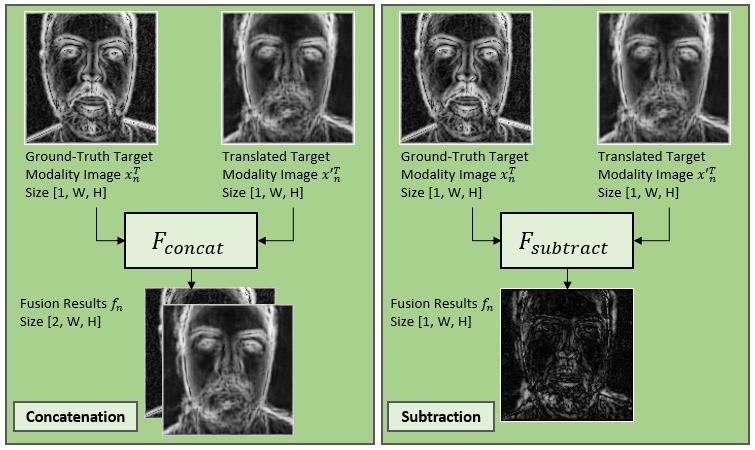}
\caption{The figure shows the illustration of two fusion operations. $F_{concat}$ and $F_{subtract}$ denote the concatenation and subtraction operation respectively. $W$ and $H$ denote the width and height of the image.}
\label{fig:figure_03}
\end{figure}
As is shown in Fig.\ref{fig:figure_02}, the translated image $x_{n}^{'T}$ is fused with the ground-truth target modality image $x_{n}^{T}$ captured by camera sensor. We have implemented two alternative operations (concatenation and subtraction) for the fusion layer. The subtraction operation aggregates information by directly computing the absolute value of the pixel-level difference. The operation result is a straightforward measurement of the similarity between two images, and we believe it's effective information for the discriminator to distinguish spoofing face images from genuine ones. The concatenation operation aggregates two images along the channel axis, which is commonly used in the general I2I task\cite{CVPR17_PIX2PIX}. The information of two images is aggregated according to the spatial position and fully kept in the operation result, which provides the discriminator detail information to learn high-level discrepancy between genuine and spoofing face images. 

The mathematical representations are shown as Eq. (\ref{eq:equation_10}) and Eq. (\ref{eq:equation_11}). $[\cdot ; \cdot]$ denotes the concatenation of two images along the channel axis. $||\cdot||$ and $-$ are pixel-wise operations that denote the calculation of absolute value and subtraction, respectively. For subtraction one, we replicate the fusion result along the channel axis to make the shape the same as the concatenation one. 
\begin{equation}
\label{eq:equation_10}
f_{n} = F_{concat}({x'}_{n}^{T},x_{n}^{T})=[{x'}_{n}^{T};x_{n}^{T}].
\end{equation}
\begin{equation}
\label{eq:equation_11}
f_{n} = F_{subtract}({x'}_{n}^{T},x_{n}^{T}) = ||{x'}_{n}^{T} - x_{n}^{T}||.
\end{equation}
Fig. \ref{fig:figure_03} illustrates the two fusion operations. Based on the experimental results in Section \ref{section:fusion-operations}, we finally selected the concatenation one as the fusion layer.

The fusion result $f_{n}$ is fed as the input to a discriminator, which outputs a patch map $t_{n}$ that can be interpreted as the patch-wise discrepancy. The output $t_{n}$ of the discriminator is supervised with binary cross-entropy (BCE) loss in a pixel-wise way. The discrimination loss is shown as
\begin{equation}
\begin{aligned}
\mathcal{L}_{dis} & = \frac{1}{N} \sum_{n=1}^{N} -(y_{n}\log(t_{n})+(1-y_{n})\log(1-t_{n})),\\
y_{n} & =\left\{
\begin{aligned}
0, &     & genuine,\\
1, &     & attack.
\end{aligned}
\right.
\end{aligned}
\label{eq:equation_12}
\end{equation}
Following \cite{ICB19_DPB}, the network architecture of the discriminator is implemented based on blocks proposed in DenseNet \cite{CVPR17_DENSENET}. More details about the architecture can be found in supplementary materials.

Under our framework, the DC module is jointly trained with the AMT module in an end-to-end way. Therefore, the total loss function is represented as 
\begin{equation}
\begin{aligned}
\mathcal{L}_{Total} & = \mathcal{L}_{AMT} + \lambda_{3} \cdot \mathcal{L}_{dis}\\
& = \lambda_{1} \cdot \mathcal{L}_{pixel} + \lambda_{2} \cdot \mathcal{L}_{latent} + \lambda_{3} \cdot \mathcal{L}_{dis}.
\end{aligned}
\label{eq:equation_13}
\end{equation}

\begin{figure}[tbp]
\centering
\subfloat[Genuine Face Images]{
\includegraphics[width=0.48\textwidth]{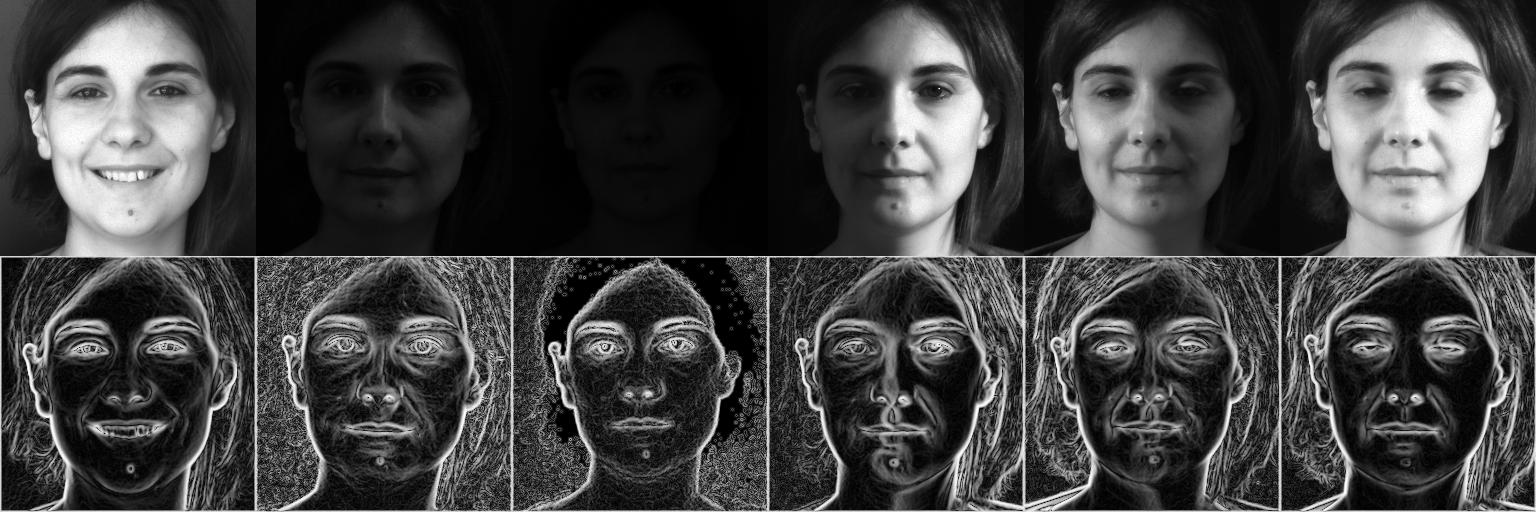}
}
\quad
\subfloat[Printed Face Images]{
\includegraphics[width=0.48\textwidth]{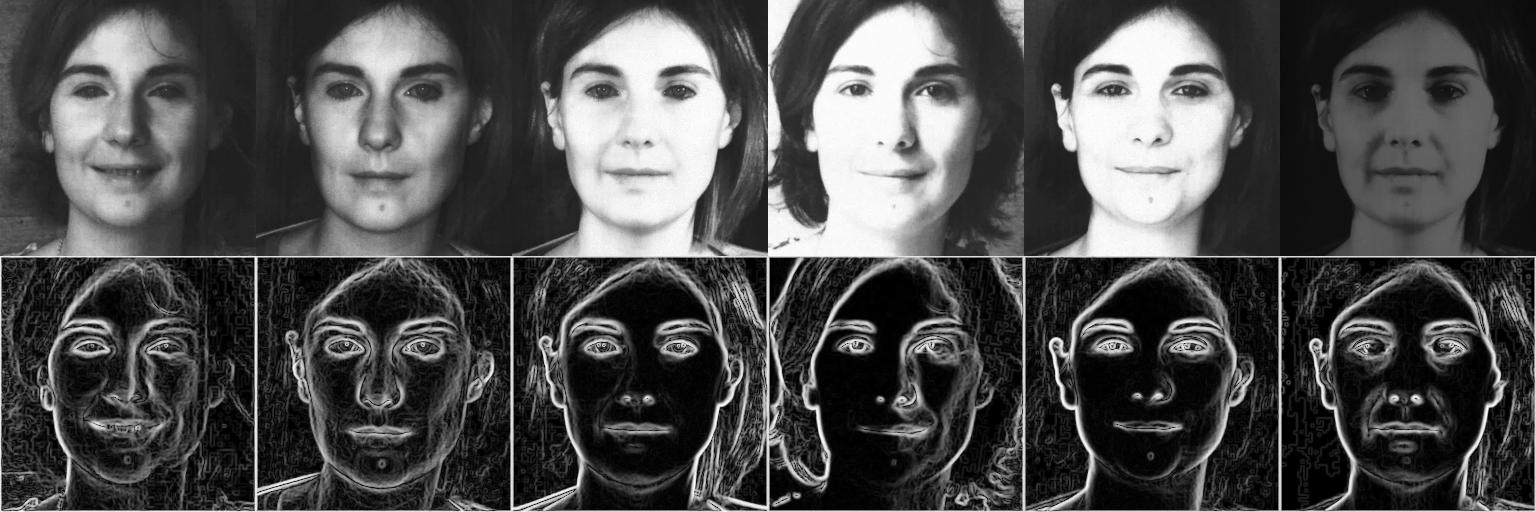}
}
\caption{The first row of (a) shows VIS images of genuine faces in MSSPOOF dataset that captured under different illumination conditions. The second row of (a) shows corresponding VIS images processed by IN module (IN-VIS). Sub-figure (b) shows VIS and IN-VIS images of attack samples}
\label{fig:figure_04}
\end{figure}

\subsection{PLGF-based Illumination Normalization}
As is known, VIS-based face PAD methods are sensitive to the variation of illumination conditions\cite{FG17_OULUNPU}. This problem also exists in bi-modality scenarios, where some modalities like VIS or NIR are employed. Such illumination variation is even tougher trouble for the training of our asymmetric modality translator, which makes the target of training unclear. For example, under different illuminations, the same image of depth modality may correspond to different VIS images of considerable intensity variation. 

To reduce the impact of illumination variation, we design an IN module based on PLGF descriptor \cite{TPAMI21_PLGF}, which is initially proposed for illumination-invariant face recognition. Images of different modalities are selectively processed by predefined convolution masks to filter the illumination components. Only images of illumination-sensitive modalities such as VIS and NIR are processed in our design, while images of other modalities skip this normalization.   
The PLGF descriptor is represented as
\begin{equation}
x_{n} = \arctan(\sqrt{(\frac{r_{n}*M_{x}}{r_{n}})^2  + (\frac{r_{n}*M_{y}}{r_{n}})^2 }).
\label{eq:equation_14}
\end{equation}
$r_{n}$ and $x_{n}$ are the raw and processed images respectively, $M_{x}$ and $M_{y}$ are two filter masks, * is convolution operation and all the other mathematical operations in Eq. (\ref{eq:equation_14}) are pixel-wise. The representations of two filter masks are shown as
\begin{equation}
M_{x}(p,q) =\left\{
\begin{aligned}
&\frac{\cos(\arctan(q/p))}{p^2 + q^2}, & (p^{2} + q^{2}) > 0,\\
&0,     & otherwise,
\end{aligned}
\right.
\label{eq:equation_15}
\end{equation}
\begin{equation}
M_{y}(p,q) =\left\{
\begin{aligned}
&\frac{\sin(\arctan(q/p))}{p^2 + q^2}, & (p^{2} + q^{2}) > 0,\\
&0,     & otherwise.
\end{aligned}
\right.
\label{eq:equation_16}
\end{equation}
$p$ and $q$ are indexes denoting the relative position to the center. In our IN module, the size of masks are set to $5$, therefore the range of $p$, $q$ is $-2 \leq p, q \leq 2$. The visualization results are shown in Fig. \ref{fig:figure_04}.

\section{Experimental Results\label{section:experimental-results}}
\subsection{Datasets Preparation}
To verify the effectiveness of our proposed method, we conducted extensive experiments on 3 publicly available datasets: Wide-Multi Channel Presentation Attack (WMCA) \cite{TIFS20_MCCNN}, CASIA-SURF \cite{CVPR19_CASIA_SURF, TBIOM_CASIA_SURF}, and Multispectral-Spoof Database (MSSPOOF) \cite{SPRINGER15_MSSPOOF}, which are commonly used for multi-modality face anti-spoofing research. The cause of choosing these datasets is that they vary in multiple aspects such as sample size, data modality, attack type, specification of the sensor device, and even preprocessing schemes. We believe this will provide a more general evaluation. The details of dataset information and the data preprocessing methods are introduced as follows.

\subsubsection{WMCA Dataset}
The WMCA dataset covers genuine faces and 7 categories of attack samples. Each original video is processed into 50 images by preprocessing procedures introduced in \cite{TIFS20_MCCNN}. In specific, there are 28540, 27550, 27850 samples in \emph{train}, \emph{dev}, \emph{test} subset respectively. Each data sample contains images of VIS (V for short), NIR (I for short), thermal (T for short), and depth (D for short) modality. Grouping each two of them, there are 6 combinations. In this work, we evaluated our method mainly under WMCA (V-I), WMCA (T-V), and WMCA (T-I) settings.

\subsubsection{CASIA-SURF Dataset}
The CASIA-SURF dataset contains genuine faces and 6 types of 2D attack samples of VIS (V for short), NIR (I for short), and depth (D for short) modality. The dataset has different versions, and we use the same one as \cite{TBIOM_CASIA_SURF} in our experiments for fair comparisons with previous work. For the data preprocessing scheme, original video clips are processed into cropped images of the face region by specific procedures as introduced in \cite{CVPR19_CASIA_SURF}. We use the processed data provided in the dataset directly for experiments, and images of V modality are converted to grayscale to keep consistent with WMCA and MSSPOOF. Eventually, there are about 49K, 16K, 98K samples in \emph{train}, \emph{dev}, \emph{test} set respectively. Each data sample has 3 images of V, I, and D modality. We pair each two of them and establish 3 bi-modality settings referred to as CASIA-SURF (V-I), CASIA-SURF (D-V), and CASIA-SURF (D-I).

\subsubsection{MSSPOOF Dataset}
The Multispectral-Spoof Database (MSSPOOF) contains paired images of VIS (V for short) and NIR (I for short) modality, which are captured under various environments. It covers genuine face images, printed VIS, and NIR images. Following the data preprocessing procedures introduced in \cite{TIFS20_MCCNN}, we align images according to the eye-center position by using landmark annotations provided in the dataset and crop face regions. After data preprocessing, there are 946, 641, 639 pairs of samples in \emph{train}, \emph{dev}, \emph{test}, and we established the modality setting denoted as MSSPOOF (V-I).
\begin{table}[tbp]
\centering
\caption{List of Dataset Information}
\resizebox{8.8cm}{!}{
\begin{tabular}{|c|c|c|}
\hline
Dataset & Modality Settings & Types of Attacks\\
\hline
\hline
\multirow{3}{*}{WMCA}
& VIS(V)-NIR(I) &  Glasses, Fake Head, Print, \\
& Thermal(T)-VIS(V) & Replay, Rigid Mask,\\
& Thermal(T)-NIR(I) & Flexible Mask, Paper Mask\\
\hline 
\multirow{6}{*}{CASIA-SURF}
& & Print(bent, eyes cropped)\\
& VIS(V)-NIR(I) & Print(bent, eyes-nose cropped)\\
& Depth(D)-VIS(V) & Print(bent, eyes-nose-mouth cropped)\\
& Depth(D)-NIR(I) & Print(still, eyes cropped)\\
& & Print(still, eyes-nose cropped)\\
& & Print(still, eyes-nose-mouth cropped)\\
\hline
MSSPOOF & VIS(V)-NIR(I) & Print(VIS image), Print(NIR image)\\
\hline
\end{tabular}
}
\label{tab:table_01}
\end{table}

The modality and type of attack information of the three datasets is listed in Tab. \ref{tab:table_01}.
\subsection{Benchmark Methods}
To verify the effectiveness of our proposed method and compare it with SOTA methods under fair experimental settings, we have also re-implemented 2 multi-modality-based face PAD methods \cite{TBIOM20_MCDPB, CVPRW20_MCCDCN} as our benchmarks. We also compared with other multi-modality-based methods \cite{CVPR19_CASIA_SURF, TBIOM_CASIA_SURF, TIFS20_MCCNN} when under comparable settings.

\subsubsection{MC-PixBiS}
MC-PixBiS\cite{TBIOM20_MCDPB} is a multi-modality face PAD method extended from DeepPixBiS \cite{ICB19_DPB}, which achieves good performance under different multi-modality face PAD benchmarks. This method concatenates images of different modalities as the input to a discriminator based on DenseNet \cite{CVPR17_DENSENET}.
The training process is end-to-end supervised by pixel-wise binary and binary labels. In our experiments, we set the parameters of this method as the same to \cite{TBIOM20_MCDPB} and denote it as MC-PixBiS.

\subsubsection{MM-CDCN}
MM-CDCN \cite{CVPRW20_MCCDCN} is based on CDC proposed in \cite{CVPR20_CDCN}, which is originally designed to address VIS-based face PAD problem. Compared to conventional convolution, CDC learns more detailed features, which is suitable for face PAD task. In our experiment, we adopt the input-level fusion like MC-PixBiS and denote it as MM-CDCN.

\subsubsection{Other Methods}
Besides aforementioned methods implemented by us, we also compared performance with other methods such as RDWT+Haralick \cite{BTAS16_HARALICK}, IQM \cite{TIP14_IQA, TIFS15_IDA}+LBP\cite{ICIP15_LBP}, MC-CNN \cite{TIFS20_MCCNN}, Single-Scale SEF \cite{CVPR19_CASIA_SURF} and Multi-Scale SEF\cite{TBIOM_CASIA_SURF} when evaluated under comparable experimental settings.

\subsection{Protocols}
\subsubsection{Grand-Test Protocol}
Grand-test is a basic protocol to verify the effectiveness of face PAD methods. It simulates an ideal situation to evaluate the overall performance. For this protocol, all the \emph{train}, \emph{dev} and \emph{test} sets contain genuine samples and all types of attack samples.

\subsubsection{Unseen-Attack Protocols}
Unseen-attack evaluation is more realistic in practical scenarios. Unlike grand-test protocol, specific attack types evaluated at the test stage are not available at the train and development stage.
Leave-one-out (LOO) protocol is commonly used for unseen-attack evaluation. For each sub-protocol, one specific type of attack is left out from the \emph{train} and \emph{dev} set, while the \emph{test} set only remains genuine samples and the left type of attack ones. For the CASIA-SURF dataset, in addition to the LOO protocol, we also evaluated our methods under the protocol used in \cite{CVPR19_CASIA_SURF, TBIOM_CASIA_SURF}, which we noted Leave-three-out (LTO). We have not conducted unseen-attack evaluations on the MSSPOOF dataset due to the small population of data samples and limited attack types.

\subsubsection{Cross-Illumination Protocol}
Cross-illumination evaluation is a necessary step to test the robustness of face PAD systems under varying illumination conditions. Similar to unseen-attack evaluation, we also adopt the LOO protocol for cross-illumination evaluation. Specifically, for each sub-protocol, we leave samples under one illumination out from the \emph{train} and \emph{dev} subsets, while in the \emph{test} set, only the samples under the left illumination condition are evaluated. We conducted cross-illumination evaluations on WMCA dataset because it covers samples under various illumination conditions and provides illumination labels.  

\subsubsection{Cross-Dataset Protocols}
Cross-dataset evaluation is the most challenging scenarios, which is used to evaluate the robustness of trained model under domain shift caused by multiple factors. V-I modality setting of the three datasets are used for cross-dataset experiments. For each dataset, we train a model on its own \emph{train} set and evaluate the \emph{the same} model on the \emph{test} sets of other two datasets, in addition to its own \emph{test} set. Therefore, there are 6 sub-protocols which are refered to as $M \xrightarrow\ W$, $W \xrightarrow\ M$, $C \xrightarrow\ W$, $W \xrightarrow\ C$, $C \xrightarrow\ M$ and $C \xrightarrow\ M$, $M \xrightarrow\ C$. $W$, $M$, and $C$ denote WMCA, MSSPOOF, and CASIA-SURF, respectively.

\subsection{Evaluation Metrics}
Various metrics have been proposed for performance evaluation in face PAD research, and different metrics are usually used in different works for performance reports. For comprehensive comparisons, we used Attack Presentation Classification Error Rate (APCER), Bonafide Presentation Classification Error Rate (BPCER), Average Classification Error Rate (ACER), Equal Error Rate (EER), TDR@FDR=1\%, and Area Under Curve (AUC) in our experiments. Larger values are better for AUC and TDR@FDR=1\%, while smaller values are better for others. 

Face PAD is a typical binary classification task, for which the threshold is usually used for decision making. AUC measures the overall performance over different thresholds. APCER, BPCER, ACER, EER, and TDR@FDR=1\% are metrics to evaluate performance at a fixed threshold. APCER measures the error rate that the system classifies attack samples as genuine ones. In complementary, BPCER measures the error rate that the system classifies genuine samples as attack ones. ACER is the average of APCER and BPCER and used to evaluate the overall performance. Following \cite{TIFS20_MCCNN}, we report APCER, BPCER, and ACER at the threshold where BPCER=1\% on \emph{dev} set. While AUC and TDR@FDR=1\% are calculated directly on \emph{test} set as in \cite{CVPR19_CASIA_SURF, TBIOM_CASIA_SURF}.

\subsection{Implementation Details}
Since the quantity of attack samples is much larger than genuine ones in all three datasets, we balanced the \emph{train} set by uniformly upsampling the genuine ones with factors 5, 4, and 3 on CASIA-SURF, WMCA, and MSSPOOF to make the amounts of genuine and attack samples in \emph{train} set comparable.
Cropped face images were resized to $128 \times 128$ before fed into the proposed bi-modality framework. The $\lambda_{1}$, $\lambda_{2}$ and $\lambda_{3}$ in Eq.(\ref{eq:equation_13}) are set as $5 \times 10^{-1}$, $1 \times 10^{-3}$ and $1$, respectively. The training process is optimized by Adam optimizer with the mini-batch size of 32 and the learning rate, which is initially set as $10^{-4}$ then decreases by half for every 10 epochs. According to the number of data samples in the training set, the model is trained for up to 20, 30, and 120 epochs on CASIA-SURF, WMCA, and MSSPOOF datasets. The proposed framework and benchmark methods are implemented based on PyTorch version 1.7.0, and all of our experiments are conducted on GPUs with CUDA version 10.1. 
\subsection{Grand-Test Evaluation}
In order to verify the effectiveness of our proposed method, we conducted extensive experiments with 7 modality settings from 3 datasets and compared the performance with SOTA methods in terms of AUC, TDR@FDR=1\%  APCER, BPCER, and ACER.
\subsubsection{Performance on WMCA}
\begin{figure*}[htbp]
\centering
\subfloat[Genuine]{
\includegraphics[width=0.12\textwidth]{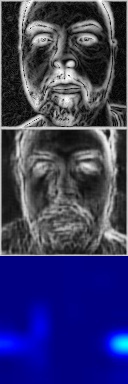}
}
\hspace{-3mm}
\subfloat[Genuine]{
\includegraphics[width=0.12\textwidth]{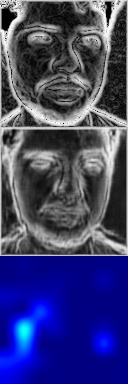}
}
\hspace{-3mm}
\subfloat[Genuine]{
\includegraphics[width=0.12\textwidth]{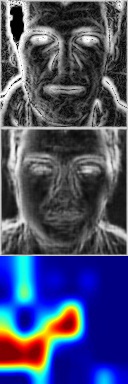}
}
\hspace{-3mm}
\subfloat[Print]{
\includegraphics[width=0.12\textwidth]{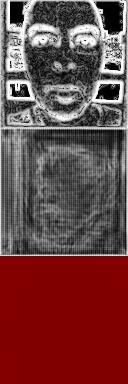}
}
\hspace{-3mm}
\subfloat[Paper]{
\includegraphics[width=0.12\textwidth]{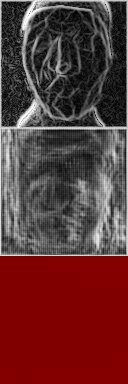}
}
\hspace{-3mm}
\subfloat[Flexible]{
\includegraphics[width=0.12\textwidth]{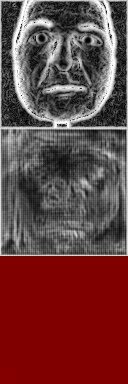}
}
\hspace{-3mm}
\subfloat[Rigid]{
\includegraphics[width=0.12\textwidth]{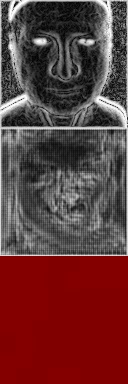}
}
\hspace{-3mm}
\subfloat[Replay]{
\includegraphics[width=0.12\textwidth]{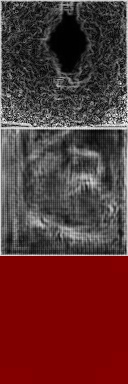}
}
\caption{Visualization of images under WMCA (T-I) setting. Columns (a)-(c) are images of genuine face samples, and columns (d)-(h) are images of attack samples. From top to bottom row are images of ground-truth IN-NIR, IN-NIR that translated from thermal images and corresponding patch maps. The patch maps are colorized as heat maps, where the red region indicates anomaly.}
\label{fig:figure_05}
\end{figure*}

We firstly compared our method with two baseline methods under the same modality settings. As is shown in Tab. \ref{tab:table_02}, MC-PixBiS has comparable performance to MM-CDCN under the T-I setting and better performance under both V-I and T-V settings. Our method clearly outperforms two baseline methods under all settings. In specific, our method reduces the ACER by $3.69\%$, $1.04\%$ and $ 2.05\%$ under V-I, T-V, and T-I setting severally, compared to MC-PixBiS. By comparing performance over different modality settings, we find that all three methods achieve their best performance under T-I settings, which indicates the importance of modality selection. Besides, we also compared with other methods which use additional modalities. From Tab. \ref{tab:table_03}, we can see that our method significantly outperforms RDWT+Haralick and IQM+LBP method under different bi-modality settings. Moreover, under the T-I setting, our method performs competitively to MC-CNN while uses fewer modalities. Some visualizations of our method are shown in Fig.\ref{fig:figure_05}.
\begin{table}[t]
\centering
\caption{Grand-Test Evaluation Results On WMCA Under The Same Modality Settings}
\resizebox{8.8cm}{!}{
\begin{tabular}{|c|l|c|c|c|c|c|}
\hline
\multirow{3}{*}{Modality Setting} & \multirow{3}{*}{Method} & \multicolumn{5}{c|}{Metrics(\%)}\\
\cline{3-7}
& & AUC & TDR@FDR=1\% & APCER & BPCER & ACER \\
\cline{3-7}  
& & ↑ & ↑ & ↓ & ↓ & ↓ \\
\hline
\hline
\multirow{3}{*}{V-I} 
& MM-CDCN\cite{CVPRW20_MCCDCN} & 97.41 & 81.70 & 17.38 & 1.15  & 9.26 \\
\cline{2-7}
& MC-PixBiS\cite{TBIOM20_MCDPB} & 99.92 & 97.64 & 9.76  & \textbf{0.00} & 4.88 \\
\cline{2-7}
& Ours & \textbf{99.94} & \textbf{98.64} & \textbf{1.41}  & 0.97  & \textbf{1.19} \\
\hline
\hline
\multirow{3}{*}{T-V} 
& MM-CDCN\cite{CVPRW20_MCCDCN} & 98.83 & 92.34 & 7.14  & 1.27  & 4.20 \\
\cline{2-7}
& MC-PixBiS\cite{TBIOM20_MCDPB} & 98.64 & 96.38 & 4.10  &\textbf{0.89} & 2.50 \\
\cline{2-7}
& Ours & \textbf{99.92} & \textbf{98.99} & \textbf{0.86} & 2.05  & \textbf{1.46} \\
\hline
\hline
\multirow{3}{*}{T-I}
& MM-CDCN\cite{CVPRW20_MCCDCN} & 99.20 & 96.05 & 3.94  & 1.08  & 2.51 \\
\cline{2-7}
& MC-PixBiS\cite{TBIOM20_MCDPB} & 99.12 & 96.34 & 3.84  & 0.71  & 2.28 \\
\cline{2-7}
& Ours & \textbf{99.99} & \textbf{99.71} & \textbf{0.45} & \textbf{0.02} & \textbf{0.23} \\
\hline
\end{tabular}
}
\label{tab:table_02}
\end{table}

\begin{table}[t]
\centering
\caption{Grand-Test Evaluation Results On WMCA Under Different Modality Settings}
\resizebox{8.8cm}{!}{
\begin{tabular}{|l|c|c|c|c|c|}
\hline
\multirow{3}{*}{Method} & \multicolumn{5}{c|}{Metrics(\%)}\\
\cline{2-6}
& AUC & TDR@FDR=1\% & APCER & BPCER & ACER \\
\cline{2-6}
& ↑ & ↑ & ↓ & ↓ & ↓ \\
\hline
\hline
RDWT+Haralick (V-I-T-D)\cite{TIFS20_MCCNN} & / & / & 6.39  & 0.49  & 3.44 \\
\hline
IQM+LBP (V-I-T-D)\cite{TIFS20_MCCNN} & / & / & 13.92 & 1.17  & 7.54 \\
\hline
MC-CNN (V-I-T-D)\cite{TIFS20_MCCNN} & / & / & 0.60  & \textbf{0.00} & 0.30 \\
\hline
\hline
Ours (V-I) & 99.94 & 98.64 & 1.41  & 0.97  & 1.19 \\
\hline
Ours (T-V) & 99.92 & 98.99 & 0.86  & 2.05  & 1.46 \\
\hline
Ours (T-I) & \textbf{99.99} & \textbf{99.71} & \textbf{0.45} & 0.02  & \textbf{0.23} \\
\hline
\end{tabular}
}
\label{tab:table_03}
\end{table}
\begin{table}[ht]
\centering
\caption{Grand-Test Evaluation Results On CASIA-SURF}
\resizebox{8.8cm}{!}{
\begin{tabular}{|c|l|c|c|c|c|c|}
\hline
\multirow{3}{*}{Modality Setting} & \multirow{3}{*}{Method} & \multicolumn{5}{c|}{Metrics(\%)}\\
\cline{3-7}
& & AUC & TDR@FDR=1\% & APCER & BPCER & ACER \\
\cline{3-7}  
& & ↑ & ↑ & ↓ & ↓ & ↓ \\
\hline
\hline
\multirow{3}{*}{V-I}
& MM-CDCN\cite{CVPRW20_MCCDCN} & 99.24 & 89.45 & 11.83 & 0.80  & 6.32 \\
\cline{2-7}
& MC-PixBiS\cite{TBIOM20_MCDPB} & 93.52 & 73.06 & 26.23 & 0.98  & 13.60 \\
\cline{2-7}
& Ours & \textbf{99.79} & \textbf{96.04} & \textbf{3.94} & \textbf{0.78}  & \textbf{2.36} \\
\hline
\hline
\multirow{3}{*}{D-V}
& MM-CDCN\cite{CVPRW20_MCCDCN} & 99.88 & 98.28 & 1.22  & 1.48  & 1.35 \\
\cline{2-7}
& MC-PixBiS\cite{TBIOM20_MCDPB} & 99.36 & 94.58 & 3.43  & 1.84  & 2.63 \\
\cline{2-7}
& Ours & \textbf{99.97} & \textbf{99.45} & \textbf{0.45} & \textbf{1.02}  & \textbf{0.74} \\
\hline
\hline
\multirow{3}{*}{D-I}
& MM-CDCN\cite{CVPRW20_MCCDCN} & 99.73 & 95.74 & 2.38  & 1.78  & 2.08 \\
\cline{2-7}
& MC-PixBiS\cite{TBIOM20_MCDPB} & 99.30 & 87.84 & 7.63  & \textbf{1.61}  & 4.62 \\
\cline{2-7}
& Ours & \textbf{99.97} & \textbf{99.66} & \textbf{0.20} & 1.73  & \textbf{0.96}\\
\hline

\end{tabular}
}
\label{tab:table_04}
\end{table}
\begin{table}[ht]
\centering
\caption{Grand-Test Evaluation Results On MSSPOOF}
\resizebox{8.8cm}{!}{
\begin{tabular}{|c|l|c|c|c|c|c|}
\hline
\multirow{3}{*}{Modality Setting} & \multirow{3}{*}{Method} & \multicolumn{5}{c|}{Metrics(\%)}\\
\cline{3-7}
& & AUC & TDR@FDR=1\% & APCER & BPCER & ACER \\
\cline{3-7}  
& & ↑ & ↑ & ↓ & ↓ & ↓ \\
\hline
\hline
\multirow{3}{*}{V-I} 
& MM-CDCN\cite{CVPRW20_MCCDCN} & 99.81 & 97.22 & 1.86  & 1.92  & 1.89 \\
\cline{2-7}
& MC-PixBiS\cite{TBIOM20_MCDPB} & 99.97 & 98.61 & \textbf{0.00}  & 2.40  & 1.20 \\
\cline{2-7}
& Ours & \textbf{100.00} & \textbf{100.00} & \textbf{0.00} & \textbf{0.96} & \textbf{0.48} \\
\hline

\end{tabular}
}
\label{tab:table_05}
\end{table}

\subsubsection{Performance on CASIA-SURF}
From experiments on CASIA-SURF, we can see that our method consistently performs better than baseline methods under all three settings as in Tab.\ref{tab:table_04}. Especially for the V-I setting, where both baseline methods perform poorly, our method outperforms them by distinct margins in terms of TDR@FDR=1\%, ACER, and APCER. Performance on CASIA-SURF shows the effectiveness of our method that under challenging settings.

\subsubsection{Performance on MSSPOOF}
To verify the effectiveness of our method on small datasets, we conducted experiments on MSSPOOF as well. According to the results in Tab. \ref{tab:table_05}, we see that our method still outperforms baseline methods. Moreover, it achieves perfect performance in terms of AUC, TDR@FDR=1\%, and APCER, which demonstrates that our method is valid even with limited training data.
\subsection{Cross-Illumination Evaluation}
In order to evaluate the robustness of our method under varied illumination conditions, we conducted cross-illumination experiments on the WMCA dataset V-I modality setting, where both modalities are sensitive to illumination variations. Data samples in the WMCA dataset were captured under 7 different illumination conditions, while there are no genuine face samples under illumination 4. Therefore with the LOO, there are only 6 sub-protocols. From Tab.\ref{tab:table_06}, we can see that our method stably performs better than two baseline methods by more than $7\%$ in terms of ACER. Although MC-PixBiS achieves comparable performance in terms of AUC, the ACER is severely degraded. It is because that the threshold of ACER is determined by \emph{dev} set and different illumination conditions of \emph{dev} and \emph{test} cause the shift of decision threshold.
\begin{table*}[htbp]
\centering
\caption{Cross-Illumination (LOO) Evaluation Results On WMCA}
\resizebox{\textwidth}{!}{
\begin{tabular}{|c|c|l|c|c|c|c|c|c|c|}
\hline
\multirow{2}{*}{Modality Setting} & \multirow{2}{*}{Metrics(\%)} & \multirow{2}{*}{Method} & \multicolumn{6}{c|}{Type of Illumination} & \multirow{2}{*}{Overall}\\
\cline{4-9}  
& & & Illumination 1 & Illumination 2 & Illumination 3 & Illumination 5 & Illumination 6 &  Illumination 7 &\\
\hline
\multirow{9}{*}{V-I} 
& \multirow{3}{*}{ACER (↓)} 
& MM-CDCN\cite{CVPRW20_MCCDCN} & 8.46 & 17.35 & 13.71 & 9.79 & 9.53 & 5.43 & 10.71±4.20 \\
\cline{3-10}
& & MC-PixBiS\cite{TBIOM20_MCDPB} & 2.88 & 18.27 & 14.78 & 5.36 & 2.71 & 4.47 & 8.08±6.71\\
\cline{3-10}
& & Ours & 0.74 & 1.75 & 0.14 & 0.47 & 1.54 & 0.08 & \textbf{0.79±0.71}\\
\cline{2-10}
&\multirow{3}{*}{AUC (↑)}
& MM-CDCN\cite{CVPRW20_MCCDCN} & 98.36 & 90.66 & 96.50 & 97.71 & 96.20 & 99.29 & 96.45±3.06\\
\cline{3-10}
&& MC-PixBiS\cite{TBIOM20_MCDPB} & 99.96 & 99.38 & 98.60 & 100.00 & 99.40 & 99.51 & 99.48±0.51 \\
\cline{3-10}
&& Ours & 99.98 & 99.85 & 100.00 & 100.00 & 99.91 & 100.00 & \textbf{99.96±0.06} \\
\cline{2-10}
& \multirow{3}{*}{TDR@FDR=1\% (↑)}
& MM-CDCN\cite{CVPRW20_MCCDCN} & 80.59 & 69.97 & 79.80 & 87.45 & 81.52 & 90.85 & 81.70±7.20 \\
\cline{3-10}
&& MC-PixBiS\cite{TBIOM20_MCDPB} & 98.59 & 93.04 & 92.38 & 100.00 & 97.21 & 92.68 & 95.65±3.36\\
\cline{3-10}
&& Ours & 99.79 & 97.22 & 100.00 & 100.00 & 98.10 & 100.00 & \textbf{99.19±1.22}\\
\hline

\end{tabular}
}
\label{tab:table_06}
\end{table*}

\subsection{Unseen-Attack Evaluation}
In practical scenarios, the deployed face PAD system may always encounter novel attacks unseen to the system designers at the model training phase. Therefore, we did experiments under unseen-attack protocols to validate the robustness of our method to zero-shot attacks.
\subsubsection{Experiments on WMCA}
\begin{table*}[tbp]
\centering
\caption{Unseen-Attack (LOO) Evaluation Results on WMCA Under the Same Modality Settings}
\resizebox{\textwidth}{!}{
\begin{tabular}{|c|c|l|c|c|c|c|c|c|c|c|}
\hline
\multirow{3}{*}{Modality Setting} & \multirow{3}{*}{Metrics(\%)} & \multirow{3}{*}{Method} & \multicolumn{7}{c|}{Type of Attack} & \multirow{3}{*}{Overall}\\
\cline{4-10}
& & & Glasses & Fake Head & Print & Replay & Rigid Mask & Flexible Mask & Paper Mask &\\
\cline{4-10}  
& & & 1 & 2 & 3 & 4 & 5 & 6 & 7 &\\
\hline
\hline
\multirow{9}{*}{V-I}
&\multirow{3}{*}{ACER (↓)} 
& MM-CDCN\cite{CVPRW20_MCCDCN} & 41.55 & 44.77 & 0.98  & 0.92  & 19.34 & 19.23 & 26.77 & 21.94±17.44 \\
\cline{3-11}
&& MC-PixBiS\cite{TBIOM20_MCDPB} & 45.45 & 48.14 & 0.02 & 0.00 & 0.87  & 19.16 & 4.44  & 16.87±21.52 \\
\cline{3-11}
&& Ours & 35.66 & 0.83 & 0.03  & 0.07  & 0.20 & 7.13 & 0.44 & \textbf{6.34±13.18} \\
\cline{2-11}
&\multirow{3}{*}{AUC (↑)} 
& MM-CDCN\cite{CVPRW20_MCCDCN} & 85.24 & 97.01 & 99.99 & 99.93 & 96.13 & 93.98 & 91.56 & 94.83±5.20 \\
\cline{3-11}
&& MC-PixBiS\cite{TBIOM20_MCDPB} & 73.90 & 91.34 & 100.00 & 100.00 & 99.99 & 96.03 & 99.85 & 94.44±9.63 \\
\cline{3-11}
&& Ours &  81.68 & 99.94 & 100.00 & 100.00 & 100.00 & 98.52 & 99.96 & \textbf{97.16±6.85} \\
\cline{2-11}
&\multirow{3}{*}{TDR@FDR=1\% (↑)} 
& MM-CDCN\cite{CVPRW20_MCCDCN} &  10.27 & 37.65 & 100.00 & 99.00 & 44.16 & 36.35 & 28.03 & 50.78±34.95 \\
\cline{3-11}
&& MC-PixBiS\cite{TBIOM20_MCDPB} & 28.36 & 10.82 & 100.00 & 100.00 & 99.51 & 49.26 & 97.38 & 69.33±38.91 \\
\cline{3-11}
&& Ours & 16.45 & 98.71 & 100.00 & 100.00 & 100.00 & 80.91 & 100.00 & \textbf{85.15±31.10} \\
\hline
\hline
\multirow{9}{*}{T-V}
&\multirow{3}{*}{ACER (↓)} 
& MM-CDCN\cite{CVPRW20_MCCDCN} & 44.73 & 0.67  & 0.37  & 0.30  & 6.20  & 2.65  & 11.59 & 9.50±16.07 \\
\cline{3-11}
&& MC-PixBiS\cite{TBIOM20_MCDPB} & 45.50 & 0.18 & 0.00 & 0.43  & 2.85  & 0.46  & 6.79  & 8.03±16.70 \\
\cline{3-11}
&& Ours & 36.84 & 0.51  & 0.43  & 0.22 & 1.33 & 0.36 & 1.57 & \textbf{5.89±13.66} \\
\cline{2-11}
&\multirow{3}{*}{AUC (↑)} 
& MM-CDCN\cite{CVPRW20_MCCDCN} & 68.42 & 99.91 & 100.00 & 100.00 & 98.82 & 99.63 & 95.61 & 94.63±11.66 \\
\cline{3-11}
&& MC-PixBiS\cite{TBIOM20_MCDPB} &  49.25 & 100.00 & 100.00 & 99.99 & 99.58 & 99.93 & 98.23 & 92.43±19.05 \\
\cline{3-11}
&& Ours &  64.89 & 99.88 & 100.00 & 100.00 & 99.96 & 100.00 & 99.83 & \textbf{94.94±13.25} \\
\cline{2-11}
&\multirow{3}{*}{TDR@FDR=1\% (↑)} 
& MM-CDCN\cite{CVPRW20_MCCDCN} &  9.55  & 99.76 & 100.00 & 100.00 & 88.73 & 95.58 & 74.00 & 81.09±32.91 \\
\cline{3-11}
&& MC-PixBiS\cite{TBIOM20_MCDPB} & 2.73  & 99.88 & 100.00 & 100.00 & 94.04 & 99.97 & 86.62 & 83.32±35.89 \\
\cline{3-11}
&& Ours & 10.82 & 100.00 & 100.00 & 100.00 & 98.78 & 99.82 & 97.46 &\textbf{86.70±33.47} \\
\hline
\hline
\multirow{9}{*}{T-I}
&\multirow{3}{*}{ACER (↓)} 
& MM-CDCN\cite{CVPRW20_MCCDCN} & 38.89 & 1.14  & 0.47  & 0.44  & 5.45  & 1.77  & 5.28  & 7.63±13.95 \\
\cline{3-11}
&& MC-PixBiS\cite{TBIOM20_MCDPB} & 39.14 & 0.62  & 0.26  & 0.53  & 1.22  & 1.10  & 3.42  & 6.61±14.38 \\
\cline{3-11}
&& Ours & 37.74 & 0.43 & 0.07 & 0.02 & 0.66 & 0.31 & 0.00 & \textbf{5.60±14.17} \\
\cline{2-11}
&\multirow{3}{*}{AUC (↑)} 
& MM-CDCN\cite{CVPRW20_MCCDCN} & 74.29 & 99.96 & 100.00 & 100.00 & 98.92 & 99.88 & 98.51 & 95.94±9.56 \\
\cline{3-11}
&& MC-PixBiS\cite{TBIOM20_MCDPB} & 49.39 & 99.97 & 100.00 & 100.00 & 99.81 & 99.97 & 99.50 & 92.66±19.08 \\
\cline{3-11}
&& Ours & 78.30 & 99.97 & 100.00 & 100.00 & 100.00 & 100.00 & 100.00 & \textbf{96.90±8.20} \\
\cline{2-11}
&\multirow{3}{*}{TDR@FDR=1\% (↑)} 
& MM-CDCN\cite{CVPRW20_MCCDCN} & 17.45 & 99.29 & 100.00 & 100.00 & 90.82 & 97.84 & 89.89 & 85.04±30.11 \\
\cline{3-11}
&& MC-PixBiS\cite{TBIOM20_MCDPB} & 18.36 & 99.88 & 100.00 & 100.00 & 90.86 & 98.74 & 89.58 & 85.35±29.88 \\
\cline{3-11}
&& Ours & 15.73 & 100.00 & 100.00 & 100.00 & 100.00 & 99.91 & 100.00 & \textbf{87.95±31.85} \\
\hline
\end{tabular}
}
\label{tab:table_07}
\end{table*}
For fair comparisons, we firstly compared our method with re-implemented baselines under three modality settings of the WMCA dataset with LOO protocol. As is shown in Tab. \ref{tab:table_07}, our method has achieved better results under all evaluation settings. Specifically, our method outperforms baseline methods by more than $10.53\%$, $2.12 \%$ and $1.01 \%$ for mean ACER; by $2.33\%$, $0.31\%$ and $0.96\%$ for mean AUC; by $15.82\%$, $3.38\%$ and $2.91\%$ for mean TDR@FDR=1\%, under V-I, T-V and T-I respectively. Besides, from the results, we see that our method performs stably over different modality settings. In contrast, the performances of baseline methods degrade severely under the V-I setting, where both modalities (VIS and NIR image) are sensitive to the environment illumination. Benefiting from IN module, our method is significantly more effective to fake head, flexible mask, and paper mask attack at unseen attack scenario, compared with baselines.
To further verify the superiority of our method, in Tab. \ref{tab:table_08}, we also compared with SOTA methods under different modality settings, even though they used additional modality information. United with different modalities, our method performs better than RDWT+Haralick, IQM+LBP, and MC-CNN. Especially, our method under the T-I setting outperforms SOTA on glasses attack and has achieved nearly perfect performance on other attacks in terms of ACER.
\subsubsection{Experiments on CASIA-SURF}
\begin{table*}[htbp]
\centering
\caption{Unseen-Attack (LOO) Evaluation Results On WMCA Under Different Modality Settings}
\resizebox{\textwidth}{!}{
\begin{tabular}{|c|l|c|c|c|c|c|c|c|c|}
\hline
\multirow{3}{*}{Metrics} & \multirow{3}{*}{Method} & \multicolumn{7}{c|}{Type of Attack} & \multirow{3}{*}{Overall}\\
\cline{3-9}
& & Glasses & Fake Head & Print & Replay & Rigid Mask & Flexible Mask & Paper Mask &\\
\cline{3-9}  
& & 1 & 2 & 3 & 4 & 5 & 6 & 7 &\\
\hline
\hline
\multirow{6}{*}{ACER (↓)} 
& RDWT+Haralick (V-I-T-D)\cite{TIFS20_MCCNN} & 48.85 & 3.16  & 0.00 & 5.77  & 7.65  & 14.05 & 2.25  & 11.68±17.01 \\
\cline{2-10}
& IQM+LBP (V-I-T-D)\cite{TIFS20_MCCNN} & 50.86 & 2.38  & 2.30  & 0.84  & 14.27 & 28.58 & 16.34 & 16.51±18.16 \\
\cline{2-10}
& MC-CNN (V-I-T-D)\cite{TIFS20_MCCNN} & 42.14 & 0.00 & 0.00 & 0.12  & 0.75  & 2.52  & 0.35  & 6.55±15.72 \\
\cline{2-10}
& Ours (V-I) & 35.66 & 0.83  & 0.03  & 0.07  & 0.20 & 7.13  & 0.44  & 6.34±13.18 \\
\cline{2-10}
& Ours (T-V) & 36.84 & 0.51  & 0.43  & 0.22  & 1.33  & 0.36  & 1.57  & 5.89±13.66 \\
\cline{2-10}
& Ours (T-I) & 37.74 & 0.43  & 0.07  & 0.02 & 0.66  & 0.31 & 0.00 & \textbf{5.60±14.17} \\
\hline

\end{tabular}
}
\label{tab:table_08}
\end{table*}

On the CASIA-SURF dataset, we evaluated our method under both LOO protocol and LTO protocol. For LOO protocol, as shown in Tab. \ref{tab:table_09}, our method outperforms two baseline methods under all sub-protocols, especially for bent photo attacks with eye-nose-mouth cropping. The mean ACER, mean AUC, and mean TDR@FDR=1\% of our method are $0.98\%$, $99.93\%$, and $98.92\%$, which are quite close to the performance under the grand-test setting. For LTO protocol, in addition to our implemented baseline methods, we also compared the proposed method with two benchmark methods on CASIA-SURF, referred to as Single-Scale SEF \cite{CVPR19_CASIA_SURF} and Multi-Scale SEF \cite{TBIOM_CASIA_SURF}, under best performing modality D-I. As shown in Tab. \ref{tab:table_10}, our method excels Single-Scale SEF and MC-PixBiS by a clear margin of about $10\%$ in TDR@FDR=1\%. Compared with Multi-Scale SEF, our method achieves comparable results.
\begin{table*}[htbp]
\centering
\caption{Unseen-Attack (LOO) Evaluation Results On CASIA-SURF}
\resizebox{\textwidth}{!}{
\begin{tabular}{|c|c|l|c|c|c|c|c|c|c|}
\hline
\multirow{3}{*}{Modality Setting} & \multirow{3}{*}{Metrics(\%)} & \multirow{3}{*}{Method} & \multicolumn{6}{c|}{Type of Attack} & \multirow{3}{*}{Overall}\\
\cline{4-9}
& & & Eyes-Still & Eyes-Bent & Eyes-Nose-Still & Eyes-Nose-Bent & Eyes-Nose-Mouth-Still & Eyes-Nose-Mouth-Bent &\\
\cline{4-9}  
& & & 1 & 2 & 3 & 4 & 5 & 6 &\\
\hline
\hline
\multirow{9}{*}{D-I} 
& \multirow{3}{*}{ACER (↓)} 
& MM-CDCN\cite{CVPRW20_MCCDCN} & 2.02  & 2.53  & 1.71  & 2.18  & 1.56  & 6.41  & 2.74±1.83 \\
\cline{3-10}
& & MC-PixBiS\cite{TBIOM20_MCDPB} & 4.43  & 6.67  & 3.23  & 4.51  & 3.14  & 7.13  & 4.85±1.69 \\
\cline{3-10}
& & Ours & 1.34 & 1.26 & 0.59 & 0.86 & 0.67 & 1.17 & \textbf{0.98±0.32} \\
\cline{2-10}
&\multirow{3}{*}{AUC (↑)}
& MM-CDCN\cite{CVPRW20_MCCDCN} & 99.78 & 99.67 & 99.81 & 99.70 & 99.81 & 98.90 & 99.61±0.35 \\
\cline{3-10}
&& MC-PixBiS\cite{TBIOM20_MCDPB} & 99.15 & 97.89 & 99.22 & 98.90 & 99.61 & 96.94 & 98.62±1.01 \\
\cline{3-10}
&& Ours & 99.92 & 99.88 & 99.97 & 99.95 & 99.97 & 99.90 & \textbf{99.93±0.04} \\
\cline{2-10}
& \multirow{3}{*}{TDR@FDR=1\% (↑)}
& MM-CDCN\cite{CVPRW20_MCCDCN} & 95.83 & 94.74 & 96.99 & 95.28 & 97.32 & 78.05 & 93.04±7.41 \\
\cline{3-10}
&& MC-PixBiS\cite{TBIOM20_MCDPB} & 79.87 & 79.27 & 90.87 & 83.83 & 91.20 & 72.18 & 82.87±7.36 \\
\cline{3-10}
&& Ours & 97.86 & 98.19 & 99.75 & 99.35 & 99.52 & 98.83 & \textbf{98.92±0.76} \\
\hline

\end{tabular}
}
\label{tab:table_09}
\end{table*}
\begin{table}[tbp]
\centering
\caption{Unseen-Attack (LTO) Evaluation Results On CASIA-SURF}
\resizebox{8.8cm}{!}{
\begin{tabular}{|c|l|c|c|c|c|c|}
\hline
\multirow{3}{*}{Modality Setting} & \multirow{3}{*}{Method} & \multicolumn{5}{c|}{Metrics(\%)}\\
\cline{3-7}
& & AUC & TDR@FDR=1\% & APCER & BPCER & ACER \\
\cline{3-7}  
& & ↑ & ↑ & ↓ & ↓ & ↓ \\
\hline
\hline
\multirow{5}{*}{D-I}
& Single-Scale SEF\cite{CVPR19_CASIA_SURF} & / & 89.7 & 1.5 & 8.4 & 4.9 \\
\cline{2-7}
& Multi-Scale SEF\cite{TBIOM_CASIA_SURF} & / & 99.4 & 2.0 & \textbf{0.3} & 1.1 \\
\cline{2-7}
& MM-CDCN\cite{CVPRW20_MCCDCN} & 99.74 & 95.46 & 2.31 & 2.15  & 2.23 \\
\cline{2-7}
& MC-PixBiS\cite{TBIOM20_MCDPB} & 98.95 & 80.56 & 6.34 & 2.79 & 4.56 \\
\cline{2-7}
& Ours & \textbf{99.95} & \textbf{99.46} & \textbf{0.54} & 0.99 & \textbf{0.77} \\
\hline

\end{tabular}
}
\label{tab:table_10}
\end{table}

\begin{table}[t]
\centering
\caption{Cross-Dataset Evaluation Results}
\resizebox{8.8cm}{!}{
\begin{tabular}{|l|c|c|c|c|c|c|c|c|}
\hline
\multirow{3}{*}{Method} 
& \multicolumn{4}{c|}{$M \xrightarrow\ W$} & \multicolumn{4}{c|}{$W \xrightarrow\ M$} \\
\cline{2-9}
& \multicolumn{2}{c|}{Intra-Performance} & \multicolumn{2}{c|}{Inter-Performance} & \multicolumn{2}{c|}{Intra-Performance} & \multicolumn{2}{c|}{Inter-Performance}\\
\cline{2-9}
& EER ↓ & AUC ↑ & EER ↓ & AUC ↑ & EER ↓ & AUC ↑ & EER ↓ & AUC ↑ \\
\hline
MM-CDCN\cite{CVPRW20_MCCDCN} & 1.89 & 99.81 & 58.41 & 39.34 & 7.51 & 97.41 & 48.05 &	54.75\\
\hline
MC-PixBiS\cite{TBIOM20_MCDPB} & 1.42 & 99.97 & 44.46 & 55.20 & 1.54 & 99.92 & 47.22 & 58.43\\
\hline
Ours & \textbf{0.47} & \textbf{100.00} & \textbf{35.88} & \textbf{69.77} & \textbf{1.15} & \textbf{99.94} & \textbf{24.08} & \textbf{86.76}\\
\hline
\hline

\multirow{3}{*}{Method} 
& \multicolumn{4}{c|}{$C \xrightarrow\ W$} & \multicolumn{4}{c|}{$W \xrightarrow\ C$} \\
\cline{2-9}
& \multicolumn{2}{c|}{Intra-Performance} & \multicolumn{2}{c|}{Inter-Performance} & \multicolumn{2}{c|}{Intra-Performance} & \multicolumn{2}{c|}{Inter-Performance}\\
\cline{2-9}
& EER ↓ & AUC ↑ & EER ↓ & AUC ↑ & EER ↓ & AUC ↑ & EER ↓ & AUC ↑ \\
\hline
MM-CDCN\cite{CVPRW20_MCCDCN} & 4.01 & 99.24 & 54.06 & 44.83 & 7.51 & 97.41 & \textbf{38.54} & \textbf{65.40} \\
\hline
MC-PixBiS\cite{TBIOM20_MCDPB} & 11.63 & 93.52 & \textbf{49.88} & \textbf{52.51} & 15.4 & 99.92 & 46.30 & 52.14 \\
\hline
Ours & \textbf{2.00} & \textbf{99.79} & 52.59 & 48.41 & \textbf{1.15} & \textbf{99.94} & 53.77 & 52.84 \\
\hline
\hline

\multirow{3}{*}{Method} 
& \multicolumn{4}{c|}{$C \xrightarrow\ M$} & \multicolumn{4}{c|}{$M \xrightarrow\ C$} \\
\cline{2-9}
& \multicolumn{2}{c|}{Intra-Performance} & \multicolumn{2}{c|}{Inter-Performance} & \multicolumn{2}{c|}{Intra-Performance} & \multicolumn{2}{c|}{Inter-Performance}\\
\cline{2-9}
& EER ↓ & AUC ↑ & EER ↓ & AUC ↑ & EER ↓ & AUC ↑ & EER ↓ & AUC ↑ \\
\hline
MM-CDCN\cite{CVPRW20_MCCDCN} & 4.01 & 99.24 & \textbf{36.01} & \textbf{72.64} & 1.89 & 99.81 & 49.40 & 51.81 \\
\hline
MC-PixBiS\cite{TBIOM20_MCDPB} & 11.63 & 93.52  & 69.19 & 32.05 & 1.42 & 99.97 & \textbf{41.60} & \textbf{61.91} \\
\hline
Ours & \textbf{2.00} & \textbf{99.79} & 42.27 & 59.49 & \textbf{0.47} & \textbf{100.00} & 44.30 & 57.44 \\
\hline
\end{tabular}
}
\label{tab:table_11}
\end{table}

\subsection{Cross-Dataset Evaluation}
We also conducted experiments under cross-dataset protocols to test the feasibility that transfers the trained model to a novel domain without fine-tuning it with newly available data. For all methods, we employed the best performing model under grand-test experiments. We chose EER, AUC as metrics for performance evaluation, and both intra-dataset and inter-dataset performance were reported based on the same model. As shown in Tab. \ref{tab:table_11}, all three evaluated methods degrade severely, and no method outstands for all sub-protocols. We observe our method consistently performs better on cross-evaluation between WMCA and MSSPOOF, compared with baseline methods. However, it does not perform well for $C-W$ and $C-M$. We believe it is because the data preprocessing conducted on CASIA-SURF is quite different from WMCA and MSSPOOF. For the $M-W$ setting, although we applied the same preprocessing procedures, the performance is still far from expected. The cause is that data samples in two datasets are captured with camera sensors of different specifications and under different environments. From our experiments, we find that the cross-dataset generalization problem on multi-modality face PAD research may not be easier than single VIS-based ones. The specifications of sensors and data preprocessing scheme should be unified to protect the performance from being severely degraded when multi-modality-based face PAD systems land for practical application.
\subsection{Ablation Study}
To better analyze our proposed method, we did series of ablation experiments to study the contribution of different components and their scalability. All the experiments are conducted under the CASIA-SURF (D-I) and WMCA (T-I) setting, where our methods show superiority compared to under other modality settings.
\begin{figure}[t]
\centering
\subfloat[Genuine]{
\includegraphics[width=0.08\textwidth]{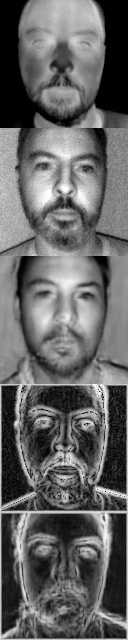}
}
\hspace{-4mm}
\subfloat[Print]{
\includegraphics[width=0.08\textwidth]{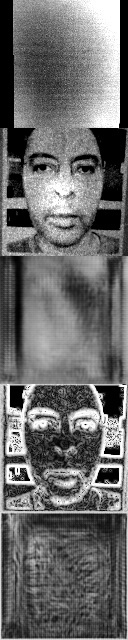}
}
\hspace{-4mm}
\subfloat[Replay]{
\includegraphics[width=0.08\textwidth]{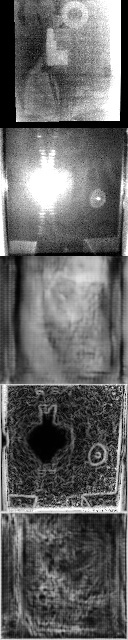}
}
\hspace{-4mm}
\subfloat[Rigid]{
\includegraphics[width=0.08\textwidth]{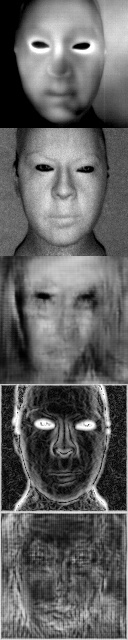}
}
\hspace{-4mm}
\subfloat[Flexible]{
\includegraphics[width=0.08\textwidth]{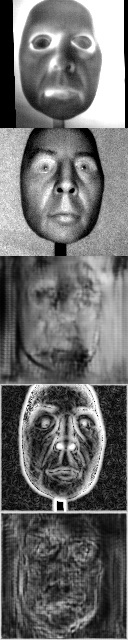}
}
\hspace{-4mm}
\subfloat[Paper]{
\includegraphics[width=0.08\textwidth]{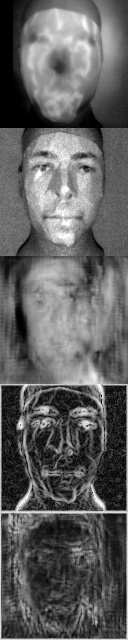}
}
\caption{Visualization of images under WMCA (T-I) setting. Column (a) are images of a same genuine face sample, and columns (b)-(f) are images of different types of attacks. From top to bottom row are images of ground-truth thermal, ground-truth NIR, NIR that translated from thermal, ground-truth IN-NIR and IN-NIR that translated from thermal images. Noted that samples of glasses attack and fake head attack are not visualized here because no samples are in authorized list due to the privacy issue.}
\label{fig:figure_06}
\end{figure}
\begin{table}[tbp]
\centering
\caption{Contribution of Different Components}
\resizebox{8.8cm}{!}{
\begin{tabular}{|c|l|c|c|c|c|c|}
\hline
\multirow{3}{*}{Modality Setting} & \multirow{3}{*}{Variants} & \multicolumn{5}{c|}{Metrics(\%)}\\
\cline{3-7}
& & AUC & TDR@FDR=1\% & APCER & BPCER & ACER \\
\cline{3-7}  
& & ↑ & ↑ & ↓ & ↓ & ↓ \\
\hline
\hline
\multirow{3}{*}{WMCA(T-I)} 
& w/o IN & 99.91 & 98.17 & 1.57  & 1.13  & 1.35 \\
\cline{2-7}
& w/o $\mathcal{L}_{AMT}$ & 99.61 & 96.06 & 3.02  & 1.77  & 2.40\\
\cline{2-7}
& full & \textbf{99.99} & \textbf{99.71} & \textbf{0.45} & \textbf{0.02} & \textbf{0.23} \\
\hline
\hline
\multirow{3}{*}{CASIA-SURF(D-I)} 
& w/o IN & 99.84 & 98.04 & 1.60  & \textbf{1.21}  & 1.41 \\
\cline{2-7}
& w/o $\mathcal{L}_{AMT}$ & 99.56 & 92.91 & 5.58  & 1.41  & 3.50 \\
\cline{2-7}
& full & \textbf{99.97} & \textbf{99.66} & \textbf{0.20} & 1.73  & \textbf{0.96} \\
\hline

\end{tabular}
}
\label{tab:table_12}
\end{table}

\subsubsection{Contribution of Different Components}
We firstly conducted module ablation tests to study the necessity of different components. As shown in Tab. \ref{tab:table_12}, both the IN module and AMT loss contribute to improving the performance under both evaluation settings. Without the IN module, the performance of our method degrades $1.54\%$, $1.12\%$ in terms of TDR@FDR=1\%, ACER under the WMCA (T-I) setting. While under the CASIA-SURF (D-I) setting, the degradation is about $1.62\%$ and $0.45\%$. Without the AMT loss, there is a degradation of $3.65\%$, $2.17\%$ and $6.75\%$, $2.54\%$ under the WMCA (T-I) and CASIA-SURF (D-I), respectively.
Fig. \ref{fig:figure_06} shows the visualization of ground-truth and translated images under WMCA (T-I) settings. Comparing the third and fifth row with the second and fourth row, we can see that the genuine face images are better translated from thermal modality to NIR modality, while attack samples fail. It shows the trained asymmetric modality translator meets our expectation. Compared to translated raw NIR images, translated NIR images which are processed by IN module (IN-NIR) between different attacks are of smaller difference.
\subsubsection{Comparison Between Different Supervision Schemes of Translator Training}
\begin{figure}[tbp]
\centering
\subfloat[Gen.]{
\includegraphics[width=0.068\textwidth]{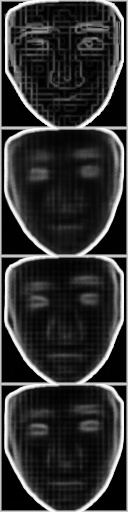}
}
\hspace{-4mm}
\subfloat[ES]{
\includegraphics[width=0.068\textwidth]{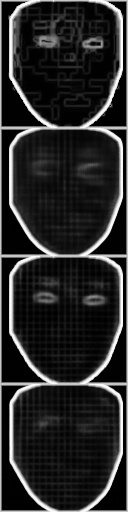}
}
\hspace{-4mm}
\subfloat[EB]{
\includegraphics[width=0.068\textwidth]{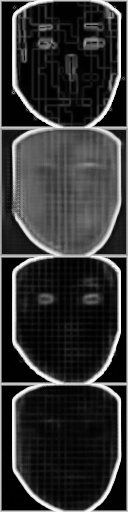}
}
\hspace{-4mm}
\subfloat[ENS]{
\includegraphics[width=0.068\textwidth]{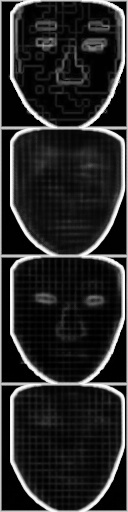}
}
\hspace{-4mm}
\subfloat[ENB]{
\includegraphics[width=0.068\textwidth]{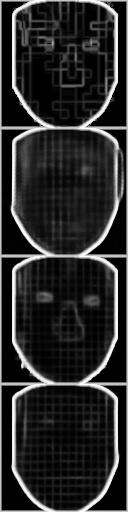}
}
\hspace{-4mm}
\subfloat[ENMS]{
\includegraphics[width=0.068\textwidth]{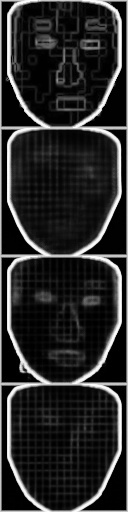}
}
\hspace{-4mm}
\subfloat[ENMB]{
\includegraphics[width=0.068\textwidth]{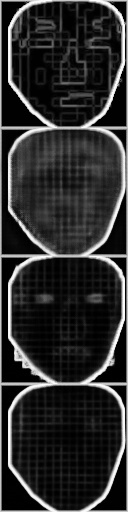}
}
\caption{Visualization of images under CASIA-SURF (D-I) setting. Column (a) are genuine face images from CASIA-SURF with different supervision schemes, and columns (b)-(g) are images of different types of attacks. From top to bottom row are ground-truth IN-NIR and IN-NIR translated by $v_{1}$, $v_{2}$ and $v_{0}$.}
\label{fig:figure_07}
\end{figure}
\begin{table}[htbp]
\centering
\caption{Comparison Between Different Supervision Schemes of Translator Training}
\resizebox{8.8cm}{!}{
\begin{tabular}{|c|l|c|c|c|c|c|}
\hline
\multirow{3}{*}{Modality Setting} & \multirow{3}{*}{Variants} & \multicolumn{5}{c|}{Metrics(\%)}\\
\cline{3-7}
& & AUC & TDR@FDR=1\% & APCER & BPCER & ACER \\
\cline{3-7}  
& & ↑ & ↑ & ↓ & ↓ & ↓ \\
\hline
\hline
\multirow{3}{*}{WMCA(T-I)} 
& $v_{1}$ & 99.74 & 97.89 & 1.70  & 1.95  & 1.82 \\
\cline{2-7}
& $v_{2}$ & 99.85 & 98.09 & 2.35  & 0.85  & 1.60 \\
\cline{2-7}
& $v_{0}$ & \textbf{99.99} & \textbf{99.71} & \textbf{0.45} & \textbf{0.02} & \textbf{0.23} \\
\hline
\hline
\multirow{3}{*}{CASIA-SURF(D-I)} 
& $v_{1}$ & 99.55 & 93.29 & 5.11  & 1.28  & 3.20 \\
\cline{2-7}
& $v_{2}$ & 99.66 & 96.50 & 3.50  & \textbf{1.08} & 2.29 \\
\cline{2-7}
& $v_{0}$ & \textbf{99.97} & \textbf{99.66} & \textbf{0.20} & 1.73  & \textbf{0.96} \\
\hline

\end{tabular}
}
\label{tab:table_13}
\end{table}

In addition to the final version of our proposed method, referred to $v_{0}$, we also implemented two variants, referred to $v_{1}$ and $v_{2}$. The only difference between the three variants is that the translator is supervised with different supervision schemes during the training process. For $v_{1}$ and $v_{2}$, the projector is removed and the $\mathcal{L}_{AMT}$ is replaced by terms $\mathcal{L}_{v_{1}} $and $\mathcal{L}_{v_{2}}$ as represented in Eq. (\ref{eq:equation_17}), where $c$ is a constant map. $\mathcal{L}_{v_{1}}$ guides the translator to translate genuine samples to target modality and attack samples to constant maps. While $\mathcal{L}_{v_{2}}$ guides the translator to translate both genuine and attack samples to the target modality.
\begin{equation}
\begin{aligned}
\mathcal{L}_{v_{1}} & = \frac{1}{N}( \sum_{n\in G}  \parallel {x'}_{n}^{T} - {x}_{n}^{T} \parallel + \sum_{n\in A} \parallel {x'}_{n}^{T} - c  \parallel),\\
\mathcal{L}_{v_{2}} & = \frac{1}{N} \sum_{n=1}^{N} \parallel {x'}_{n}^{T} - x_{n}^T  \parallel.
\end{aligned}
\label{eq:equation_17}
\end{equation}
The comparison is conducted in both numerical and visual. We can see from Tab. \ref{tab:table_13}, under WMCA (T-I) setting, $v_{1}$ performs comparable with $v_{2}$, and the $v_{0}$ slightly outperforms them. And under CASIA-SURF (D-I) setting, the $v_{1}$ performs the worst, and the $v_{0}$ performs the best. Comparing $v_{0}$ to $v_{1}$, there is $6.37\%$ improvement in terms of TDR@FDR=1\% and $2.24\%$ improvement in terms of ACER. The visualization results are shown in Fig. \ref{fig:figure_07}. We can see that for all three variants, genuine face images can be successfully translated. While for the translation of attack images, results of $v_{1}$ are different from the genuine one, but the variance between different attacks is obvious; results of $v_{2}$ are similar to the ground truth, and for some attack types the results are indistinguishable from genuine ones; results of the final $v_{0}$ are different from the genuine ones, and show consistency over different attacks.
\begin{figure}[t]
\centering
\subfloat[Genuine]{
\includegraphics[width=0.08\textwidth]{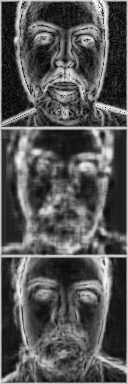}
}
\hspace{-4mm}
\subfloat[Print]{
\includegraphics[width=0.08\textwidth]{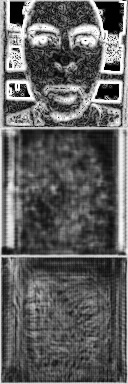}
}
\hspace{-4mm}
\subfloat[Replay]{
\includegraphics[width=0.08\textwidth]{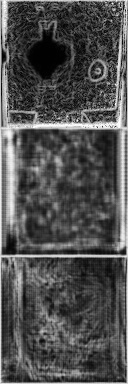}
}
\hspace{-4mm}
\subfloat[Rigid]{
\includegraphics[width=0.08\textwidth]{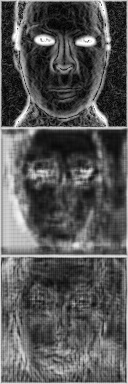}
}
\hspace{-4mm}
\subfloat[Flexible]{
\includegraphics[width=0.08\textwidth]{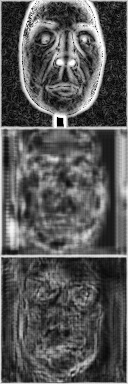}
}
\hspace{-4mm}
\subfloat[Paper]{
\includegraphics[width=0.08\textwidth]{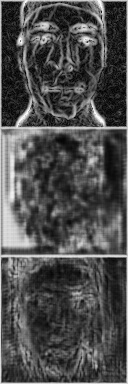}
}
\caption{Visualization of images under WMCA (T-I) setting. Column (a) are images of a same genuine face sample from WMCA, and columns (b)-(f) are images of different attacks. From top to bottom row are images of ground-truth IN-NIR, translated IN-NIR without and with the translation block.}
\label{fig:figure_08}
\end{figure}
\begin{table}[t]
\centering
\caption{Comparison Between Different Fusion Operations}
\resizebox{8.8cm}{!}{
\begin{tabular}{|c|l|c|c|c|c|c|}
\hline
\multirow{3}{*}{Modality Setting} & \multirow{3}{*}{Variants} & \multicolumn{5}{c|}{Metrics(\%)}\\
\cline{3-7}
& & AUC & TDR@FDR=1\% & APCER & BPCER & ACER \\
\cline{3-7}  
& & ↑ & ↑ & ↓ & ↓ & ↓ \\
\hline
\hline
\multirow{2}{*}{WMCA(T-I)} 
& subtraction & 99.46 & 97.07 & 2.85  & 1.25  & 2.05 \\
\cline{2-7}
& concatenation & \textbf{99.99} & \textbf{99.71} & \textbf{0.45} & \textbf{0.02} & \textbf{0.23} \\
\hline
\hline
\multirow{2}{*}{CASIA-SURF(D-I)} 
& subtraction & 99.92 & 98.56 & 1.11  & \textbf{1.24}  & 1.18 \\
\cline{2-7}
& concatenation & \textbf{99.97} & \textbf{99.66} & \textbf{0.20} & 1.73  & \textbf{0.96} \\
\hline

\end{tabular}
}
\label{tab:table_14}
\end{table}
\subsubsection{Comparison Between Different Fusion Operations \label{section:fusion-operations} }
We compared two operations for fusion, computing the difference of translated images with the ground truth and concatenating them. As shown in Tab. \ref{tab:table_14}, under CASIA-SURF (D-I), both fusion operations perform well, and there is no obvious difference; while under the WMCA (T-I) setting, the concatenation one stands out. Therefore, we experimentally chose the concatenation one for our framework.
\subsubsection{Sensitivity to the Absence of the Translation Block}
Following the architecture used in general I2I translation tasks, we use a translation block in our translator architecture. Therefore, we also conducted experiments to study the sensitivity of our method to the absence of the translation block. From the visualization results in Fig. \ref{fig:figure_08}, we find that there is a noticeable quality degradation in visual without the block. Both genuine face and attack images become blurred. However, as in Tab. \ref{tab:table_15}, there is no severe decline in the numerical performance. Since the translation is only an auxiliary task to improve the final attack detection accuracy, our experimental results indicate the potential for architecture pruning to further improve efficiency if necessary.
\subsubsection{Compatibility of Illumination Normalization Module With Existing Methods}
Since the proposed IN module works as a preprocessing procedure and can be easily added to other multi-modality face PAD methods, we finally test the compatibility of IN module with two baseline methods. From Tab. \ref{tab:table_16}, we find that the PLGF-based illumination normalization also improves the performance of the MC-PixBiS method, especially under CASIA-SURF (D-I) evaluation setting. While there is no obvious improvement for MM-CDCN, we believe it is because the CDC used in MM-CDCN itself possesses kernel-wised normalization functionality.
\begin{table}[t]
\centering
\caption{Necessity of The Translation Block}
\resizebox{8.8cm}{!}{
\begin{tabular}{|c|l|c|c|c|c|c|}
\hline
\multirow{3}{*}{Modality Setting} & \multirow{3}{*}{Variants} & \multicolumn{5}{c|}{Metrics(\%)}\\
\cline{3-7}
& & AUC & TDR@FDR=1\% & APCER & BPCER & ACER \\
\cline{3-7}  
& & ↑ & ↑ & ↓ & ↓ & ↓ \\
\hline
\hline
\multirow{3}{*}{WMCA(T-I)} 
& w/o TB & 99.94 & 99.52 & 0.55  & 0.68  & 0.61 \\
\cline{2-7}
& with TB & \textbf{99.99} & \textbf{99.71} & \textbf{0.45} & \textbf{0.02} & \textbf{0.23} \\
\hline
\hline
\multirow{3}{*}{CASIA-SURF(D-I)} 
& w/o TB  & 99.90 & 98.25 & 1.75  & \textbf{0.76} & 1.25 \\
\cline{2-7}
& with TB & \textbf{99.97} & \textbf{99.66} & \textbf{0.20} & 1.73  & \textbf{0.96} \\
\hline

\end{tabular}
}
\label{tab:table_15}
\end{table}

\begin{table}[H]
\centering
\caption{Effectiveness of Illumination Normalization Module}
\resizebox{8.8cm}{!}{
\begin{tabular}{|c|l|c|c|c|c|c|}
\hline
\multirow{3}{*}{Modality Setting} & \multirow{3}{*}{Variants} & \multicolumn{5}{c|}{Metrics(\%)}\\
\cline{3-7}
& & AUC & TDR@FDR=1\% & APCER & BPCER & ACER \\
\cline{3-7}  
&  & ↑ & ↑ & ↓ & ↓ & ↓ \\
\hline
\hline
\multirow{3}{*}{WMCA(T-I)} 
& MC-PixBiS & 99.12 & 96.34 & 3.84  & \textbf{0.71} & 2.28 \\
\cline{2-7}
& MC-PixBiS + IN & \textbf{99.22} & \textbf{97.69} & \textbf{2.33} & 0.94  & \textbf{1.63} \\
\hline
\hline
\multirow{3}{*}{CASIA-SURF(D-I)} 
& MC-PixBiS & 99.30 & 87.84 & 7.63  & 1.61  & 4.62 \\
\cline{2-7}
& MC-PixBiS + IN & \textbf{99.78} & \textbf{97.31} & \textbf{2.02} & \textbf{1.60} & \textbf{1.81} \\
\hline
\hline
\multirow{3}{*}{WMCA(T-I)} 
& MM-CDCN & \textbf{99.20} & 96.05 & 3.94  & \textbf{1.08}  & 2.51 \\
\cline{2-7}
& MM-CDCN + IN & 99.16 & \textbf{96.80} & \textbf{2.58} & 1.55 & \textbf{2.06} \\
\hline
\hline
\multirow{3}{*}{CASIA-SURF(D-I)} 
& MM-CDCN & 99.73 & 95.74 & \textbf{2.38}  & 1.78  & 2.08 \\
\cline{2-7}
& MM-CDCN + IN & \textbf{99.80} & \textbf{96.63} & \textbf{2.38} & \textbf{1.71} & \textbf{2.04} \\
\hline

\end{tabular}
}
\label{tab:table_16}
\end{table}
\subsection{Discussions}
\subsubsection{Cross-Domain Evaluation Performance}
According to the experimental results, our method achieves good performance in grand-test and cross-illumination evaluation. However, the performance of cross-dataset evaluation is not satisfactory. Besides, the performance for the detection of unseen attacks such as disguising glasses attack could be further improved.

\subsubsection{Extension to K-Modality Settings}
Although our method is proposed to address the face PAD problem under bi-modality scenarios, we attempted to extend it to K(K$>$2) modality scenarios by performing score fusion with different bi-modality models. We did some experiments on WMCA and CASIA-SURF dataset in 3 modality scenario, and the results are shown in Tab. \ref{tab:table_17} and \ref{tab:table_18} respectively. According to our experimental results, the AUC, TDR@FDR=1\%, and APCER can be further improved with the score fusion of different bi-modality models. However, naively extending the proposed method with multiple model score fusion will multiply the model size and the computational cost. More efficient schemes for extension should be explored in future work.
\subsubsection{Data Augmentations}
The good performance of deep learning-based face PAD methods relies on sufficient and diverse training data. However, it's usually hard to collect enough training data to cover the test domain in practical applications. Data augmentation techniques such as brightness, contrast, saturation, and hue adjustment have been used for single VIS-modality-based face PAD method to improve the accuracy and generalization performance. But most of the data augmentation techniques for face PAD methods are specifically designed for VIS images and can not be directly applied to other modalities like depth maps and thermal images. The exploration of data augmentation techniques for multi-modality data is a promising direction to further improve the generalization performance of multi-modality-based face PAD methods.
\begin{table}[H]
\centering
\caption{Score Fusion Results On WMCA Grand-Test Evaluation}
\resizebox{8.8cm}{!}{
\begin{tabular}{|c|c|c|c|c|c|}
\hline
\multirow{3}{*}{Method} & \multicolumn{5}{c|}{Metrics(\%)}\\
\cline{2-6}
& AUC & TDR@FDR=1\% & APCER & BPCER & ACER \\
\cline{2-6}
& ↑ & ↑ & ↓ & ↓ & ↓ \\
\hline
\hline
T-V & 99.919 & 98.99 & 0.86 & 2.05 & 1.46 \\
\hline
T-I & 99.989 & 99.71 & 0.45 & \textbf{0.02} & \textbf{0.23}\\
\hline
Score Fusion & \textbf{99.995} & \textbf{99.84} & \textbf{0.30} & 0.17 & \textbf{0.23} \\
\hline
\end{tabular}
}
\label{tab:table_17}
\end{table}
\begin{table}[H]
\centering
\caption{Score Fusion Results On CASIA-SURF Grand-Test Evaluation}
\resizebox{8.8cm}{!}{
\begin{tabular}{|c|c|c|c|c|c|}
\hline
\multirow{3}{*}{Method} & \multicolumn{5}{c|}{Metrics(\%)}\\
\cline{2-6}
& AUC & TDR@FDR=1\% & APCER & BPCER & ACER \\
\cline{2-6}
& ↑ & ↑ & ↓ & ↓ & ↓ \\
\hline
\hline
D-V & 99.965 & 99.45 & 0.45 & \textbf{1.02} & \textbf{0.74} \\
\hline
D-I & 99.968 & 99.66 & 0.20 & 1.73 & 0.96 \\
\hline
Score Fusion & \textbf{99.988} & \textbf{99.91} & \textbf{0.05} & 1.79 & 0.92 \\
\hline
\end{tabular}
}
\label{tab:table_18}
\end{table}
\subsubsection{Different Evaluation Metrics}
\begin{figure}[t]
\centering
\includegraphics[width=0.48\textwidth]{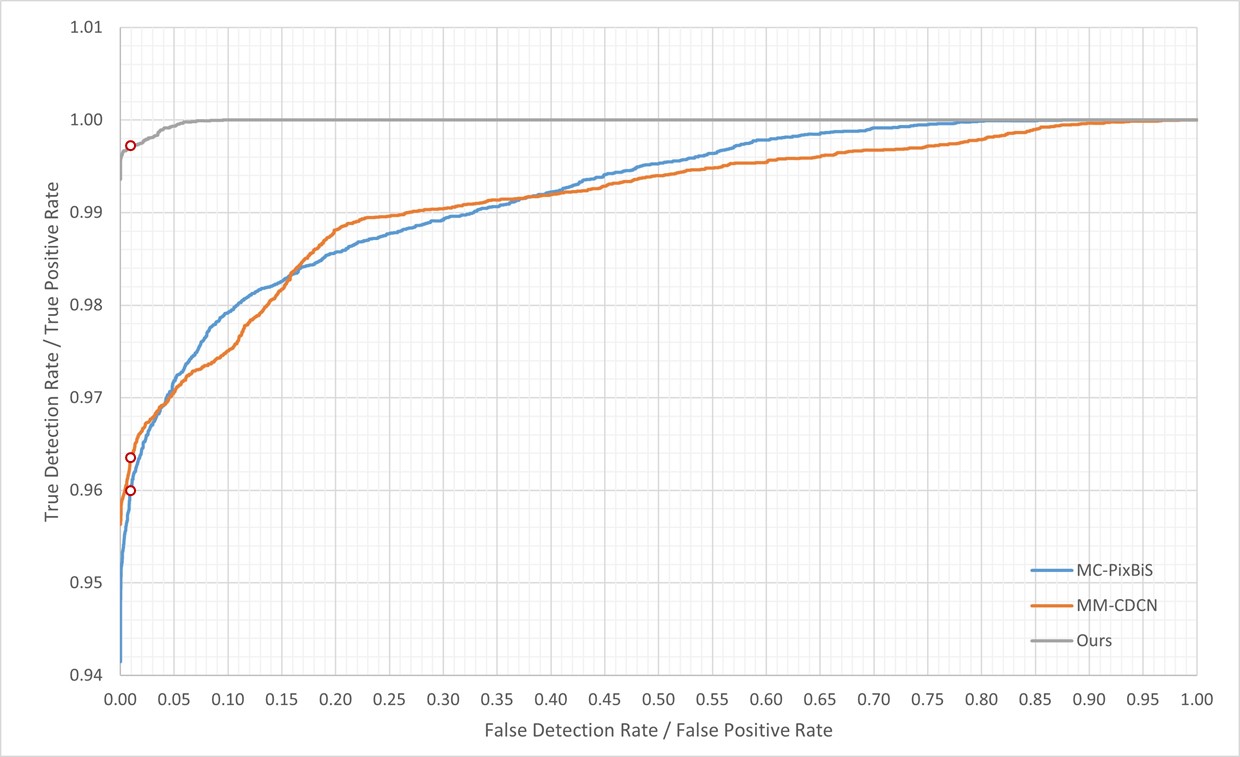}
\caption{Visualization of Receiver Operating Characteristic (ROC) curves for WMCA(T-I) grand-test evaluation. The horizontal axis represents the False Detection Rate (FDR) and the vertical axis represents the True Detection Rate (TDR).}
\label{fig:figure_09}
\end{figure}

In our experiments, we evaluated the performance with different metrics such as AUC, TDR@FDR=1\%, APCER, BPCER, and ACER. By visualizing the ROC curves as in Fig.\ref{fig:figure_09}, we found that both benchmark methods and our method achieve close to perfect TDRs at the thresholds of higher FDRs, but our method shows significant superiority with lower FDRs. AUC measures the overall performance at the thresholds of different FDRs. While metrics TDR@FDR=1\%, APCER, BPCER, and ACER measure the performance at fixed thresholds of lower FDR only. Therefore, the performance improvement of our method in terms of AUC is not as large as other metrics.

\subsubsection{Comparison Between Different Modalities}
To verify the effectiveness of our proposed method, we did experiments under different bi-modality settings of WMCA and CASIA-SURF datasets. On the WMCA dataset, our method with T-I modality outperforms V-I and T-V modality. On the CASIA-SURF dataset, our method with D-V and D-I modality performs better than V-I modality, and the difference between D-V modality and D-I modality is not significant.
\section{Conclusions}
In this work, we propose a bi-modality framework for face presentation attack detection based on asymmetric modality translation. Under the framework, the connection between two modality images of genuine faces is established via an asymmetric modality translator and used as the cue for spoofing attack detection. We did extensive experiments on several public datasets to verify the effectiveness of our proposed method. The experimental results show that our method applies to different modality settings and can effectively detect various seen and unseen attacks under varying illumination conditions. To further improve the proposed method, data augmentation techniques for multi-modality data and efficient extension schemes for additional modalities could be explored in future work.

\bibliographystyle{IEEEtran}
\bibliography{body/reference.bib}

\newpage
\markboth{Supplementary Materials}{}

\setcounter{table}{18}
\setcounter{section}{5}

\section{Implementation Details of Asymmetric Modality Translation Module}
\begin{table}[h]
\centering
\caption{Implementation Details of Encoder (0.36M Params)}
\resizebox{8.8cm}{!}{
\begin{tabular}{|c|l|r|r|}
\hline
Block & Layer & Kernel Shape & Output Shape\\
\hline
\hline
\multirow{3}{*}{1} & Conv2d & [1, 64, 7, 7] & [1, 64, 128, 128]\\
\cline{2-4}
& InstanceNorm2d & / & [1, 64, 128, 128]\\
\cline{2-4}
& ReLU & / & [1, 64, 128, 128] \\
\hline
\hline
\multirow{3}{*}{2} & Conv2d & [64, 128, 3, 3] & [1, 128, 64, 64]\\
\cline{2-4}
& InstanceNorm2d & / & [1, 128, 64, 64]\\
\cline{2-4}
& ReLU & / & [1, 128, 64, 64] \\
\hline
\hline
\multirow{3}{*}{3} & Conv2d & [128, 256, 3, 3] & [1, 256, 32, 32]\\
\cline{2-4}
& InstanceNorm2d & / & [1, 256, 32, 32]\\
\cline{2-4}
& ReLU & / & [1, 256, 32, 32] \\
\hline
\end{tabular}
}
\label{tab:table_19}
\end{table}
\begin{table}[h]
\centering
\caption{Implementation Details of Decoder (0.36M Params)}
\resizebox{8.8cm}{!}{
\begin{tabular}{|c|l|r|r|}
\hline
Block & Layer & Kernel Shape & Output Shape\\
\hline
\hline
\multirow{3}{*}{1} & ConvTranspose2d & [128, 256, 3, 3] & [1, 128, 64, 64]\\
\cline{2-4}
& InstanceNorm2d & / & [1, 128, 64, 64]\\
\cline{2-4}
& ReLU & / & [1, 128, 64, 64] \\
\hline
\hline
\multirow{3}{*}{2} & ConvTranspose2d & [64, 128, 3, 3] & [1, 64, 128, 128]\\
\cline{2-4}
& InstanceNorm2d & / & [1, 64, 128, 128]\\
\cline{2-4}
& ReLU & / & [1, 64, 128, 128] \\
\hline
\hline
\multirow{2}{*}{3} & Conv2d & [64, 1, 7, 7] & [1, 1, 128, 128]\\
\cline{2-4}
& Tanh & / & [1, 1, 128, 128] \\
\hline
\end{tabular}
}
\label{tab:table_20}
\end{table}
\begin{table}[h]
\centering
\caption{Implementation Details of Translation Blocks (10.62M Params)}
\resizebox{8.8cm}{!}{
\begin{tabular}{|c|l|r|r|}
\hline
Block & Layer & Kernel Shape & Output Shape\\
\hline
\hline
\multirow{5}{*}{1-9} & Conv2d & [256, 256, 3, 3] & [1, 256, 32, 32]\\
\cline{2-4}
& InstanceNorm2d & / & [1, 256, 32, 32]\\
\cline{2-4}
& ReLU & / & [1, 256, 32, 32] \\
\cline{2-4}
& Conv2d & [256, 256, 3, 3] & [1, 256, 32, 32]\\
\cline{2-4}
& InstanceNorm2d & / & [1, 256, 32, 32] \\
\hline
\end{tabular}
}
\label{tab:table_21}
\end{table}
\begin{table}[H]
\centering
\caption{Implementation Details of Projector (0.27M Params)}
\resizebox{8.8cm}{!}{
\begin{tabular}{|c|l|r|r|}
\hline
Block & Layer & Kernel Shape & Output Shape\\
\hline
\hline
\multirow{3}{*}{1} & Conv2d & [256, 64, 3, 3] & [1, 64, 32, 32]\\
\cline{2-4}
& InstanceNorm2d & / & [1, 64, 32, 32]\\
\cline{2-4}
& ReLU & / & [1, 64, 32, 32] \\
\hline
\hline
\multirow{4}{*}{2} & Conv2d & [64, 1, 1, 1] & [1, 1, 32, 32]\\
\cline{2-4}
& InstanceNorm2d & / & [1, 1, 32, 32]\\
\cline{2-4}
& ReLU & / & [1, 1, 32, 32] \\
\cline{2-4}
& FC & [1024, 128] & [1, 128]\\
\hline
\end{tabular}
}
\label{tab:table_22}
\end{table}

\newpage
As introduced in the main content of this paper, our Asymmetric Modality Translation (AMT) Module consists of Encoder, Translation Blocks, Decoder, and Projector. The implementation details of these network architectures are presented in Tab. \ref{tab:table_19}, \ref{tab:table_20}, \ref{tab:table_21} and \ref{tab:table_22}.

\section{Implementation Details of Discrimination Module}
As introduced in main paper,
Tab. \ref{tab:table_23} shows the implementation details of the Discriminator.
\begin{table}[htbp]
\centering
\caption{Implementation Details of Discriminator (3.19M Params)}
\resizebox{8.8cm}{!}{
\begin{tabular}{|c|l|r|r|}
\hline
Block & Layer & Kernel Shape & Output Shape\\
\hline
\hline
\multirow{4}{*}{1} & Conv2d & [2, 96, 7, 7] & [1, 96, 64, 64]\\
\cline{2-4}
& BatchNorm2d & [96] & [1, 96, 64, 64]\\
\cline{2-4}
& ReLU & / & [1, 96, 64, 64] \\
\cline{2-4}
& MaxPool2D & / & [1, 96, 32, 32]\\
\hline
\hline
2 & DenseNet161.feature4 & * & [1, 48, 32, 32]\\
\hline
\hline
3 & DenseNet161.feature5 & * & [1, 192, 16, 16]\\
\hline
\hline
4 & DenseNet161.feature6 & * & [1, 48, 16, 16]\\
\hline
\hline
5 & DenseNet161.feature7 & * & [1, 384, 8, 8]\\
\hline
\hline
6 & Conv2d & [384, 1, 1, 1] & [1, 1, 8, 8]\\
\hline
\end{tabular}
}
\label{tab:table_23}
\end{table}
\section{Supplementary Experimental Results}
\subsection{Comparison about Swapping the Source and Target Modality}
We also did some comparison experiments to study the difference by swapping the source and the target modality. The comparison results are shown in Tab. \ref{tab:table_24}.
\begin{table}[htbp]
\centering
\caption{Comparison about Swapping the Source and Target Modality ON WMCA}
\resizebox{8.8cm}{!}{
\begin{tabular}{|c|l|c|c|c|c|c|}
\hline
\multirow{3}{*}{Modality Setting} & \multirow{3}{*}{Method} & \multicolumn{5}{c|}{Metrics(\%)}\\
\cline{3-7}
& & AUC & TDR@FDR=1\% & APCER & BPCER & ACER \\
\cline{3-7}  
& & ↑ & ↑ & ↓ & ↓ & ↓ \\
\hline
\hline
\multirow{2}{*}{V-I} 
& I2V & 99.85 & 98.20 & 2.20 & \textbf{0.97}  & 1.58 \\
\cline{2-7}
& V2I & \textbf{99.94} & \textbf{98.64} & \textbf{1.41}  & \textbf{0.97}  & \textbf{1.19} \\
\hline
\hline
\multirow{2}{*}{T-V} 
& V2T & 99.39 & 97.35 & 2.53  & \textbf{1.11}  & 1.82 \\
\cline{2-7}
& T2V & \textbf{99.92} & \textbf{98.99} & \textbf{0.86} & 2.05  & \textbf{1.46} \\
\hline
\hline
\multirow{2}{*}{T-I}
& T2I & 99.78 & 98.26 & 1.15 & 1.76 & 1.46 \\
\cline{2-7}
& I2T & \textbf{99.99} & \textbf{99.71} & \textbf{0.45} & \textbf{0.02} & \textbf{0.23} \\
\hline
\end{tabular}
}
\label{tab:table_24}
\end{table}
\newpage
\begin{table*}[htbp]
\centering
\caption{Unseen-Attack (LOO) Experimental Results On WMCA}
\resizebox{\textwidth}{!}{
\begin{tabular}{|c|c|c|c|c|c|c|c|c|c|}
\hline
\multirow{3}{*}{Metrics(\%)} & \multirow{3}{*}{Modality Settings} & \multicolumn{7}{c|}{Type of Attack} & \multirow{3}{*}{Overall}\\
\cline{3-9}
& & Glasses & Fake Head & Print & Replay & Rigid Mask & Flexible Mask & Paper Mask &\\
\cline{3-9}  
& & 1 & 2 & 3 & 4 & 5 & 6 & 7&\\
\hline
\hline
\multirow{6}{*}{ACER (↓)} 
& D-V & 32.09 & 0.51 & 0.61 & 0.69 & 1.86 & 15.14 & 2.40 & 7.61±12.00\\
\cline{2-10}
& D-I & 39.90 & 0.62 & 0.04 & 0.00 & 0.13 & 12.16 & 0.00 & 7.55±14.95\\
\cline{2-10}
& T-D & 41.31 & 0.00 & 0.15 & 0.00 & 0.92 & 2.17 & 1.30 & 6.55±15.40\\
\cline{2-10}
& V-I & 35.66 & 0.83  & 0.03  & 0.07  & 0.20 & 7.13  & 0.44  & 6.34±13.18\\
\cline{2-10}
& T-V & 36.84 & 0.51  & 0.43  & 0.22  & 1.33  & 0.36  & 1.57  & 5.89±13.66\\
\cline{2-10}
& T-I & 37.74 & 0.43  & 0.07  & 0.02 & 0.66  & 0.31 & 0.00 & \textbf{5.60±14.17}\\
\hline
\end{tabular}
}
\label{tab:table_25}
\end{table*}
\begin{table*}[htbp]
\centering
\caption{Unseen-Attack (LOO) Experimental Results On CASIA-SURF}
\resizebox{\textwidth}{!}{
\begin{tabular}{|c|c|c|c|c|c|c|c|c|}
\hline
\multirow{3}{*}{Metrics(\%)} & \multirow{3}{*}{Modality Settings} & \multicolumn{6}{c|}{Type of Attack} & \multirow{3}{*}{Overall}\\
\cline{3-8}
&  & Eyes-Still & Eyes-Bent & Eyes-Nose-Still & Eyes-Nose-Bent & Eyes-Nose-Mouth-Still & Eyes-Nose-Mouth-Bent &\\
\cline{3-8}  
& & 1 & 2 & 3 & 4 & 5 & 6 &\\
\hline
\hline
\multirow{3}{*}{ACER (↓)} 
& V-I & 6.82 & 5.26 & 3.30 & 2.74 & 1.45 & 12.03 & 5.27±3.82 \\
\cline{2-9}
& D-V & 2.18  & 1.24 & 0.70 & 0.41 & 0.55 & 0.97 & 1.01±0.65\\
\cline{2-9}
& D-I & 1.34 & 1.26 & 0.59 & 0.86 & 0.67 & 1.17 & \textbf{0.98±0.32} \\
\hline
\hline
\multirow{3}{*}{AUC (↑)} 
& V-I & 98.98 & 99.34 & 99.62 & 99.73 & 99.86 & 97.71 & 99.21±0.80 \\
\cline{2-9}
& D-V & 99.66 & 99.84 & 99.96 & 99.98 & 99.99 & 99.92 & 99.89±0.13\\
\cline{2-9}
& D-I & 99.92 & 99.88 & 99.97 & 99.95 & 99.97 & 99.90 & \textbf{99.93±0.04}\\
\hline
\hline
\multirow{3}{*}{TDR@FDR=1\% (↑)} 
& V-I & 86.08 & 90.46 & 94.22 & 95.60 & 97.87 & 78.90 & 90.52±7.05\\
\cline{2-9}
& D-V & 96.54 & 98.62 & 99.59 & 99.79 & 99.94 & 98.98 & 98.91±1.27\\
\cline{2-9}
& D-I & 97.86 & 98.19 & 99.75 & 99.35 & 99.52 & 98.83 & \textbf{98.92±0.76}\\
\hline
\end{tabular}
}
\label{tab:table_26}
\end{table*}
\begin{table}[H]
\centering
\caption{Glossary of Abbreviations \& Acronyms}
\resizebox{9cm}{!}{
\begin{tabular}{cc}
\hline
Abbreviations \& Acronyms & Terms\\
\hline
ACER & Average Classification Error Rate\\
AMT & Asymmetric Modality Translation\\
APCER & Attack Presentation Classification Error Rate\\
AUC & Area Under Curve\\
BCE & Binary Cross-Entropy\\
BPCER & Bonafide Presentation Classification Error Rate\\
CDC & Central Difference Convolution\\
CNN & Convolutional Neural Network\\
CUDA & Compute Unified Device Architecture\\
D & depth modality\\
DC & Discrimination\\
DL & deep learning\\
EB & Eyes-Bent\\
EER & Equal Error Rate\\
ENMB & Eyes-Nose-Mouth-Bent\\
ENMS & Eyes-Nose-Mouth-Still\\
ENB & Eyes-Nose-Bent\\
ENS & Eyes-Nose-Still\\
ES & Eyes-Still\\
FDR & False Detection Rate\\
Gen. & Genuine\\
GPU & Graphics Processing Unit\\
I & near infrared modality\\
I2I & image-to-image\\
IN & Illumination Normalization\\
IN-VIS & VIS image processed by IN module\\
IRM & Illumination-Reflectance Model\\
K & thousand\\
LOO & leave-one-out\\
LTO & leave-three-out\\
NAS & network architecture search\\
NIR & near infrared\\
PAD & presentation attack detection\\
PLGF & Pattern of Local Gravitational Force\\
ReLU & Rectified Linear Unit\\
SE & Squeeze-and-Excitation\\
SOTA & state-of-the-art\\
SWIR & short wave infrared\\
T & thermal modality\\
TDR & True Detection Rate\\
UV & ultraviolet\\
V & visible light modality\\
VIS & visible light\\
\hline
\end{tabular}
}
\label{tab:table_27}
\end{table}
\subsection{Comparison Between Different Modalities under Unseen-Attack Protocol}
The performance of our method under different modality settings on WMCA and CASIA-SURF datasets are shown in Tab. \ref{tab:table_25} and \ref{tab:table_26} respectively. On WMCA dataset, our method with T-I modality performs the best. On CASIA-SURF dataset, our method with D-V and D-I modality performs better than V-I modality and the difference between D-V modality and D-I modality is not significant.

\section{Glossary of Abbreviations \& Acronyms}
For better readability, we listed the abbreviations and acronyms of terms in Tab. \ref{tab:table_27}.
\end{document}